%% file: J-STFormer.tex
\documentclass[10pt,journal,cspaper,compsoc]{IEEEtran}
\usepackage[breaklinks=true,colorlinks,citecolor=blue,linkcolor=blue,urlcolor=blue]{hyperref}
\usepackage{url}            % simple URL typesetting
\usepackage{booktabs}       % professional-quality tables
\usepackage{amsfonts}       % blackboard math symbols
\usepackage{nicefrac}       % compact symbols for 1/2, etc.
\usepackage{times}
\usepackage{graphicx}
\usepackage{amsmath}
\usepackage{amssymb}
\usepackage{color}
\usepackage{xcolor}
\usepackage{multirow}
\usepackage{overpic}
\usepackage{amssymb}
\usepackage{pifont}
\usepackage{makecell}% to use \Xhline{1pt}
\usepackage[caption=false,font=small,labelfont=rm,textfont=rm]{subfig}

\usepackage[nocompress]{cite} %IEEE Computer Society needs nocompress option
\usepackage{ragged2e}
\usepackage{cancel}  

\usepackage{geometry}
\usepackage{silence}
\graphicspath{{./images/}}

\def\eg{\emph{e.g.,~}}
\def\ie{\emph{i.e.,~}}

\newcommand{\ourMthd}{DeST}
\newcommand{\figref}[1]{Fig.~\ref{#1}}
\newcommand{\tabref}[1]{Tab.~\ref{#1}}
\newcommand{\secref}[1]{Section~\ref{#1}}
\newcommand{\eqcref}[1]{Eq.~\ref{#1}}
\newcommand{\myPara}[1]{\vspace{.05in} \noindent\textbf{#1}}

\begin{document}

\title{A Decoupled Spatio-Temporal Framework for Skeleton-based Action Segmentation}

\author{Yun-Heng Li, Zhong-Yu Li, Shanghua Gao, Qilong Wang, Qibin Hou,~\IEEEmembership{Member,~IEEE}, \\Ming-Ming Cheng,~\IEEEmembership{Senior Member,~IEEE}

\IEEEcompsocitemizethanks{
\IEEEcompsocthanksitem Y.-H. Li, Z.-Y. Li, S. Gao, Q. Hou, and M.-M. Cheng are with VCIP, School of Computer Science, Nankai University, Tianjin 300350, China.
E-mail:(\{yunhengli; lizhongyu\}@mail.nankai.edu.cn)
(Corresponding author: Q. Hou)
\IEEEcompsocthanksitem Q. Wang is with the College of Intelligence and
Computing, Tianjin University, Tianjin 300350, China.
\IEEEcompsocthanksitem This work was supported in part by the Supercomputing Center of Nankai University.
\IEEEcompsocthanksitem Our code is available at: https://github.com/lyhisme/\ourMthd.
}
}

%%%%%%%%% ABSTRACT
\IEEEtitleabstractindextext{
\begin{abstract} \justifying
Effectively modeling discriminative spatio-temporal information is essential for segmenting activities in long action sequences. 
However, we observe that existing methods are limited in weak spatio-temporal modeling capability due to two forms of decoupled modeling: 
(i) cascaded interaction couples spatial and temporal modeling, which over-smooths motion modeling over the long sequence, and (ii) joint-shared temporal modeling adopts shared weights to model each joint, ignoring the distinct motion patterns of different joints. 
In this paper, we present a \textbf{De}coupled \textbf{S}patio-\textbf{T}emporal Framework~(\textbf{\ourMthd}) to address the above issues. 
Firstly, we decouple the cascaded spatio-temporal interaction to avoid stacking multiple spatio-temporal blocks, while achieving sufficient spatio-temporal interaction. 
Specifically, \ourMthd~performs once unified spatial modeling and divides the spatial features into different groups of sub-features, which then adaptively interact with temporal features from different layers.
Since the different sub-features contain distinct spatial semantics, the model could learn better interaction patterns at each layer.
Meanwhile, inspired by the fact that different joints move at different speeds, we propose joint-decoupled temporal modeling, which employs independent trainable weights to capture distinctive temporal features of each joint. 
On four large-scale benchmarks of different scenes, \ourMthd~significantly outperforms current state-of-the-art methods with less computational complexity. 
\end{abstract}
}

\maketitle
\IEEEdisplaynontitleabstractindextext
\IEEEpeerreviewmaketitle
%%%%%%%%% BODY TEXT

\IEEEraisesectionheading{\section{Introduction}\label{sec:introduction}}

\IEEEPARstart{T}{emporal} action segmentation (TAS), which aims at classifying each video frame, is critical for various practical applications, such as surveillance~\cite{luo2019capturing,son2019detection}, assisted rehabilitation~\cite{filtjens2020data,kidzinski2019automatic}, interactive robotics~\cite{kenney2009interactive,siam2019video} and virtual reality~\cite{sudha2017approaches,biswas2014kinect}.
Current works on action segmentation can be divided into RGB video-based methods and skeleton-based methods.
% 
% \gsh{When introducing RGB/skeleton-based methods, you need to briefly introduce the key features of these methods so people can understand the advantages/disadvantages you mentioned.}
RGB video-based approaches adopt two stages to model spatial and temporal information, respectively, where high computational costs are required to capture the pixel-level appearance information.
In contrast, skeleton-based action segmentation utilizes one stage for unified spatial and temporal modeling and is more efficient as the skeleton acquired by motion capture systems already represents motion information.
Moreover, skeleton data is more robust to appearance variation and environmental noises,~\eg scene and illumination changes~\cite{liu2017enhanced,song2022constructing,cheng2021extremely}.
Based on the skeleton data, previous works~\cite{carreira2017quo,simonyan2014two,wang2018appearance,wang2015action,wang2016temporal,liu2022video,arnab2021vivit,Wang_2023_CVPR} have achieved great progress in identifying short action clips.

\input{figs/Introduction}

Despite the remarkable progress, segmenting all actions in a long skeleton sequence is still extremely challenging.
Existing temporal action segmentation methods~\cite{filtjens2022skeleton,xu2023efficient,liu2022spatial,li2023involving} generally employ spatial and temporal graph convolution networks (GCNs) to model the spatial and temporal dependencies among joints and perform spatio-temporal interaction to extract motion features, as illustrated in~\figref{Introduction}(a). 
However, we observe that these GCNs-based approaches are limited to weak spatio-temporal modeling capability for long skeleton sequences due to two forms of coupled modeling,~\ie (i) cascaded interaction pattern and (ii) joint-shared temporal modeling.
Specifically, identifying complex actions needs to extract discriminative spatio-temporal representations by interacting with spatial and temporal information of joints.
Nevertheless, to perform multiple spatio-temporal interactions, the cascaded interaction methods alternately stack spatial and temporal modeling modules, which over-smooths the spatio-temporal information~\cite{fagcn2021,pang2023skeleton} and leads to high computational costs \cite{song2022constructing,cheng2021extremely}.
In addition, in previous temporal GCNs, joint-shared weights~\cite{filtjens2022skeleton,xu2023efficient} are used to model the temporal dependencies of all joints. 
% ~\eg 2D convolution~\cite{filtjens2022skeleton,xu2023efficient}.
% \gsh{2D convolution is the example for what?}
% 
However, this joint-shared strategy ignores the fact that different joints have different motion speeds.
For instance, as shown in \figref{skeleton-Introduction}, when a person moves, the feet typically move faster than other joints, whereas the waist moves more slowly.
Once joint-shared weights are trained on all joints, the dynamic temporal information of the foot is weakened, making it ambiguous to distinguish some actions, like ``Running'' and ``Walking.''
Thus, such joint-shared temporal modeling will lose information about joint motion differences, restricting the temporal modeling capability.

To address the limitations mentioned above, we propose a Decoupled Spatio-Temporal Framework (\ourMthd), as illustrated in~\figref{Introduction}(b). 
Unlike existing methods~\cite{filtjens2022skeleton,xu2023efficient,liu2022spatial,li2023involving}~(in~\figref{Introduction}(a)), \ourMthd~utilizes decoupled interaction (instead of cascaded interaction) to perform spatio-temporal modeling and adopts independent weights (not joint-shared weights) to model the temporal dependencies of each joint.
\input{figs/skeleton-Introduction}
Specifically, our \ourMthd~involves of two core components, including decoupled spatio-temporal interaction (DSTI) and joint-decoupled temporal modeling (JTM). 
To avoid the effect of over-smoothing brought by the coupled spatio-temporal modeling, DSTI decouples cascaded spatio-temporal interaction by adopting a single-layer unified spatial modeling and dividing the spatial features into different groups of spatial sub-features. 
Particularly, various spatial sub-features focus on distinct spatial semantics and adaptively interact with the temporal features from different layers, enabling learning better interaction patterns at different layers. 
As such, our \ourMthd~can capture richer spatio-temporal representations in long skeleton action sequences.
% 
% To generate more effective temporal features, our \ourMthd~presents a joint-decoupled temporal modeling (JTM) instead of the previous joint-shared weights strategy. 
% 
% In particular, our JTM adopts independent trainable weights to capture temporal dependencies for each joint separately. 
To generate more effective temporal features, JTM adopts independent trainable weights to capture temporal dependencies for each joint separately. 
By adopting JTM, the learned temporal dependencies can be adapted to the unique motion attributes of each joint and hence improve the temporal modeling ability of the model. 

To evaluate the effectiveness of our \ourMthd~method, experiments are conducted on four long-sequence skeleton-based TAS benchmarks with diverse scenes, including competitive sports (\eg MCFS-22~\cite{liu2021temporal} and MCFS-130~\cite{liu2021temporal} datasets), daily activities (\eg PKU-MMD~\cite{liu2017pku} dataset), and typical warehousing activities (\eg LARa~\cite{niemann2020lara} dataset). 
Our \ourMthd~consistently achieves superior performance with lower computational costs.

Our contributions can be summarized as follows:
\begin{itemize}
  \item In this paper, we propose a Decoupled Spatio-Temporal (\ourMthd) framework, which to our best knowledge, makes the first attempt to decouple spatio-temporal modeling for effective skeleton-based action segmentation. 
  \item We propose a decoupled spatio-temporal interaction 
  module and a joint-decoupled temporal modeling module to decouple the cascaded spatio-temporal interaction and learn the discriminative motion pattern of each joint, respectively. 
  \item Extensive experimental results demonstrate that our proposed \ourMthd~method achieves new state-of-the-art results with even lower model complexity than previous works.
\end{itemize}

\section{Related work}

\subsection{RGB-based Temporal Action Segmentation}

Most works follow a two-stage process,~\ie feature extracting and temporal modeling, to segment RGB-based videos.
The first stage simply uses pre-trained networks~\cite{Carreira_2017_CVPR} to extract the spatial features of each video frame.
Most action segmentation methods focus on the second stage to model temporal context information.
Early approaches~\cite{ding2017tricornet,Singh_2016_CVPR} mostly adopt recurrent neural networks for temporal modeling.
To capture long-range dependencies, temporal convolution networks are proposed for action segmentation, including encoder-decoder architecture~\cite{lea2017temporal}, temporal series pyramid networks~\cite{li2021efficient}, and multi-layer dilated convolutions~\cite{farha2019ms,li2020ms}.
Meanwhile, for these TCN-based methods, Gao et al.~\cite{Gao_2021_CVPR,gao2022rf} proposed a global-to-local search scheme to efficiently search for receptive field combinations. 
On the other hand, transformer-based models~\cite{chinayi_ASformer,behrmann2022unified,ltc2023bahrami} are proposed for action segmentation.
For example, Bahrami et al.~\cite{ltc2023bahrami} presented LTContext that leverages sparse attention to capture the long-term context and windowed attention to model the local information.
However, the above methods still tend to over-segment actions.
To address this problem, many researchers explore different approaches, \eg additional network branches~\cite{wang2020boundary,ishikawa2021alleviating}, two-step methods~\cite{li2021efficient}, or hierarchical refiner~\cite{ahn2021refining}.

Recently, multi-modal learning~\cite{li2022bridge} and diffusion models~\cite{liu2023diffusion} have also been employed for action segmentation.
However, these RGB-based methods require high-cost computational resources to extract motion representations by processing RGB images or temporal optical flow~\cite{chen2021multi}.
The generated motion representations also lose fine-grained information, especially joint connections~\cite{wang2020learning}. 
In this work, we focus on another line of methods,~\ie skeleton-based action segmentation.
Compared to RGB-based methods, skeleton-based methods are more efficient and robust to the variations of the environment in that the human skeleton conveys high-level motion information.

\input{figs/Framework}

\subsection{Skeleton-based Temporal Action Segmentation}
By using low-cost MoCap systems, the motion of the skeleton can be captured as a time series.
Filtjens et al.~\cite{filtjens2022skeleton} first proposed multi-stage spatio-temporal graph convolution neural networks (MS-GCN) by replacing the initial stage of temporal convolutions with spatio-temporal graph convolutions.
On top of MS-GCN, Xu et al.~\cite{xu2023efficient} created a combined network by connectionist temporal classification loss.
Moreover, Tian et al.~\cite{tian2023stga} proposed spatio-temporal attention blocks to model spatial correlation and temporal correlation separately.
However, these works utilize shared weights of temporal graph convolution or sliding window attention to model temporal dependencies, limiting the temporal learning capability.
To enhance temporal semantics, Liu et al.~\cite{liu2022spatial} proposed spatial focus attention to model the spatial dependencies and multi-layer TCNs to model temporal dependencies.
Similarly, Li et al.~\cite{li2023involving} proposed an involving distinguished temporal graph convolution network (IDT-GCN) to improve the capability to model spatial-temporal information and utilize extra information,~\ie the order of action proceeding, to capture the segment-level encoding features and action boundary representations. 
Nevertheless, these GCN-based methods conduct single or multiple cascaded spatio-temporal interactions, which over-smooth the spatio-temporal features and fail to capture complex spatio-temporal information effectively. 
Unlike these methods, we propose a decoupled spatio-temporal architecture that adopts unified spatial modeling to extract spatial sub-features and let these sub-features interact with temporal features to avoid cascaded spatio-temporal interactions. 
Meanwhile, we utilize independent weights to capture distinct temporal dependencies of each joint, enhancing the temporal modeling capabilities.

\subsection{Skeleton-based spatio-temporal Networks}

The spatio-temporal graph convolution networks and transformers have also been widely used to model human action dynamics for other skeleton-based action understanding tasks, including short action recognition~\cite{hao2021hypergraph,li2023graph,hu2019joint,zhang2019view}, temporal action detection~\cite{nowozin2012action,sharaf2015real,wang2018beyond,korban2023semantics}, and human motion prediction~\cite{cui2020learning,Li_2020_CVPR,Dang_2021_ICCV,li2021symbiotic}. 
However, these methods are not suitable for segmenting actions in long sequences. 
On one hand, the problems of joint-shared temporal modeling and cascaded spatio-temporal interaction, which are demonstrated in \secref{sec:introduction}, also exist in these tasks. 
For example, action recognition methods model the relations between body joints by alternately stacking spatial graph convolutions and temporal graph convolutions~\cite{yan2018spatial,Shi_2019_CVPR,cheng2020decoupling,Chen_2021_ICCV,song2020richly,kong2021symmetrical,zhu2022multilevel,10113233,xu2023spatiotemporal}, thus over-smoothing spatio-temporal information and failing to identify multiple actions in long action sequences. 
Moreover, these tasks usually model spatial and temporal information in different domains. 
For instance, some human motion prediction approaches~\cite{li2021symbiotic,aksan2021spatio,yu2023towards} separate spatio-temporal information into the spatial and temporal branches. 
Differently, the proposed \ourMthd~models the motion of joints by unified spatio-temporal modeling. 
% through joint skeleton data without utilizing multi-stream fusion strategies, aiming at performing unified spatio-temporal modeling.

\section{Method}
Let ${\mathcal V} = \left\{ {\alpha_{vt}} \mid v \in [1, V], t\in [1,  T] \right\}$ be a skeleton sequence, where $V$ and $T$ denote the number of joints and the number of frames, respectively.
Action segmentation aims to identify the category for each frame.
In this section, we propose the Decoupled spatio-temporal model~(\ourMthd), an end-to-end framework for skeleton-based action segmentation.
The overall architecture is shown in \figref{Framework}, which consists of two core components: 
1) decoupled spatio-temporal interaction~(DSTI) architecture to avoid the over-smoothing issue brought by the coupled spatio-temporal modeling and achieve sufficient spatio-temporal interaction and 2) joint-decoupled temporal modeling~(JTM) module to enhance the discrimination of temporal features.
For convenience, some important symbols and their definitions are listed in \tabref{Symbols}.

\input{tabs/I-Symbols}

\subsection{Preliminary}
\label{sec:SSM}

To make the paper understandable, we first briefly present unified spatial modeling for skeleton-based action segmentation.

\myPara{Multi-scale spatial modeling.}
Recently, graph convolution networks have been successfully adopted to aggregate spatial information of joints~\cite{wang2020learning,shi2020skeleton,zhang2019graph,cao2018skeleton,li2021symbiotic,yang2021feedback}.
Inspired by MS-G3D~\cite{Liu_2020_CVPR}, we adopt multi-scale spatial modeling to capture the spatial dependencies between joints.
Specifically, we first define a $k$-adjacency matrix $A^{(k)} \in {{\mathbb R}^{V \times V}}$ , which connects the joints at distance $k$, as follows: 
\begin{equation}
  \begin{aligned}
    A^{(k)}_{ij} = \begin{cases}1, & \text { if } d\left(\alpha_{i}, 
      \alpha_{j}\right) = k, \\ 
      1, & \text { if } i=j, \\ 
      0, & \text { otherwise, }\end{cases}
  \end{aligned}
\label{eq2}
\end{equation}
where $d\left(\alpha_{i}, \alpha_{j}\right)$ denotes the shortest distance~(\ie the number of edges) between joints $\alpha_i$ and $\alpha_j$. 
Given the $C$-dim features $X \in {{\mathbb R}^{C \times T \times V}}$ of $V$ joints, through matrix multiplication,~\ie $X A^{(k)}$, the dependencies between joints at the distance $k$ are captured.
In addition to the adjacency matrix $A^{(k)}$ defined by the physical connections of body joints, we construct a $V \times V$ adjacency matrix $B^{(k)}$ that is trainable and hence can adaptively learn the relationships between joints.
Through combining the two adjacency matrices at different distances, the multi-scale spatial features $S \in {\mathbb R^{C^{s} \times T \times V}}$ can be aggregated as follows: 
\begin{equation}
  {S} = {\rm MLP}({W^{\mathcal S} X} ([(\hat A^{(1)}+B^{(1)}) \parallel \cdots \parallel (\hat A^{(K)}+B^{(K)})])),
\label{eq3}
\end{equation}
where $\parallel\cdot\parallel$ is the concatenation operation along the second dimension, $W^{\mathcal S}$ is a weight tensor, and $\hat A^{(k)}$ is the normalized matrix\footnote{The normalized matrix can be represented as $\hat A^{(k)}=D^{{^{ - \frac{1}{2}}}}A^{(k)}D^{\frac{1}{2}}$, where $D_i = \sum_j {A_{ij}} + \beta$ is a degree matrix and $\beta$ is set to 0.001.} following~\cite{yan2018spatial,Liu_2020_CVPR}. 
The MLP means a multilayer perceptron to adjust the number of channels.
In particular, the aggregated spatial feature $S$ encodes multi-scale spatial representations by adopting multiple adjacency matrices at distances from 1 to $K$. 
By default, $K$ is set as 13 here.
% $k$ $\in$ $[1, K]$ is the distance between joints. 
% % 
% We set the default value as $K=13$.
% % 
% By adopting multiple adjacency matrices at different distances simultaneously, the aggregated spatial feature $S$ encodes multi-scale spatial representations.

\input{figs/Transform}

\subsection{Decoupled spatio-temporal Interaction}
\label{sec:STI}
It is crucial to establish the spatio-temporal interaction in skeleton-based action segmentation.
Following GCN-based methods \cite{yan2018spatial}, existing action segmentation methods~\cite{filtjens2022skeleton,xu2023efficient,liu2022spatial,li2023involving} perform cascaded spatio-temporal interaction by coupling the spatial and temporal modeling.
However, we observe that 1) some methods~\cite{filtjens2022skeleton,xu2023efficient} stack multiple cascaded spatio-temporal modeling blocks, over-smoothing motion information and producing weak spatio-temporal features.
2) Other methods~\cite{liu2022spatial,li2023involving} employ a single cascaded spatio-temporal interaction, resulting in an insufficient spatio-temporal interaction.
To solve these problems, we propose a Decoupled Spatio-Temporal Interaction (DSTI) architecture to avoid cascaded interaction and achieve efficient spatio-temporal interaction, as shown in~\figref{Framework}.

\myPara{Decoupled spatio-temporal interaction architecture.}
Specifically, we perform unified spatial modeling (introduced in \secref{sec:SSM}) only once at the beginning of the network, and make the generated spatial features interact with temporal features from different layers. 
In this way, we can achieve sufficient spatio-temporal interaction without alternately stacking spatial and temporal modeling modules. 
Meanwhile, considering that the temporal modeling modules at different layers require to interact with different spatial features, we divide the spatial features $S$ into $M$ groups along the channel dimension, producing spatial sub-features $\{S_i\in \mathbb R^{\frac{C^{s}}{M}\times T \times V} \mid i \in [1, M] \}$, as shown in~\figref{Transform}~(left). 
To enable the interaction with the temporal features, we further transform the sub-features into $\hat{S}_i \in \mathbb{R}^{T \times V} = \mathcal{R}(S_i)$ to align with the number of dimensions of temporal features, as shown in \figref{Transform}~(right). 
$\mathcal{R}$ is the transformation function and we use convolution by default. 
Finally, different groups of spatial sub-features can express distinct spatial semantics and we let them interact with different layers of temporal modeling to achieve sufficient interaction, as shown in~\figref{Framework}, while avoiding the over-smoothing of spatio-temporal semantics
caused by cascaded interaction. 

\myPara{Adaptive interaction.}
Instead of simple summation, we achieve adaptive spatio-temporal interaction by cross-attention~\cite{vaswani2017attention}.
Given the temporal features $H_{(l)}$ output by the $l$-th JTM and the spatial sub-features $\hat{S}_{(l+1)}$, the spatio-temporal features $R_{(l)} \in {\mathbb R^{{C^t} \times T}}$ can be generated as follows:
\begin{equation}
\begin{split}
    R_{(l)} &= {\rm DSTI}(\hat{S}_{l+1}, H_{(l)}) \\
    &= {\rm Softmax}\left(\frac{(W^C \hat{S}_{l+1})H_{(l)}^T}{\tau}\right) H_{(l)} + H_{(l)},
\end{split}
\label{eq8}
\end{equation} 
where $W^C$ is a 1D convolution that adjusts the dimension of the $\hat S_{l+1}$ to match that of $H_{(l)}$, and $\tau$ is the temperature parameter controlling the sharpness of the attention maps.
This flexible interaction enables our \ourMthd~to capture rich spatio-temporal features. 

\myPara{Discussion.} 
Some methods~\cite{gao2019res2net,chen2021multi,Dang_2021_ICCV} also divide features within a layer and capture multi-scale features by hierarchical connections. 
Unlike them, our \ourMthd~divides distinct spatial sub-features to adaptively interact with temporal features from different layers and aims at decoupling cascaded spatio-temporal interaction.
Experiments also show that the divided spatial sub-features can adapt to different actions and enable adaptive spatio-temporal interaction. 
Please refer to \secref{sec:exp-DSTI} for more details.
Meanwhile, compared to the multi-stream fusion methods~\cite{li2021symbiotic,aksan2021spatio,yu2023towards,song2022constructing} that decompose the spatial and temporal information into multiple separate branches, \ourMthd~utilizes a single stream to capture spatio-temporal dependencies under a unified spatio-temporal domain.

\subsection{Joint-decoupled Temporal Modeling}
\label{sec:JTM}
It is critical to capture rich temporal dependencies of joints in long skeleton sequences. 
Recent action segmentation methods~\cite{filtjens2022skeleton,xu2023efficient} perform temporal modeling over joints using 2D temporal convolutions.
Among these approaches, a major problem is that they usually capture temporal information of each joint by joint-shared weights, eliminating the differences in motion speed between joints and 
limiting the temporal representations of motion. 
On the contrast, we propose a Joint-decoupled Temporal Modeling (JTM) module that decouples the temporal modeling of different joints, \ie the temporal dependencies of each joint are captured by an independent weight, as shown in~\figref{JTM}. 
In this way, the model can capture temporal dependencies that represent the distinct motion patterns of each joint and generate rich temporal semantics.

\myPara{Temporal modeling by independent weights.}
To model the temporal dependencies of different joints separately, we employ independent joint-decoupled weights to perform the convolution operation with each joint, as shown in~\figref{JTM}. 
The input of the JTM can be spatial sub-features or spatio-temporal features generated by the spatio-temporal interaction~(DSTI) layers. 
Without loss of generality, we represent the input as $J \in {\mathbb R^{V \times T}}$. 
Taking the $v$-th $J_{v}^{} \in {\mathbb R^{1 \times T}}$ as an example, an independent convolution with kernel size $C^f$ and weights $W^J_v \in {\mathbb R^{{C^t} \times {C^f}}}$ is used to model its temporal dependencies $W^J_v \LARGE{\ast} J_{v}^{}$.
Here, $C^t$ is the number of channels and $\LARGE{\ast}$ denotes the convolution operation. 
Furthermore, by fusing the temporal dependencies of all joints, the high-level temporal features,~\ie $\mathcal T(J) \in {\mathbb R^{{C^t} \times T}}$, can be generated as follows:
\begin{equation}
  \begin{split}
  \mathcal T(J) = \sum\limits_{v=1}^V {(W^J_{v} \LARGE{\ast} J_v^{})}.
  \end{split}
\label{eq5}
\end{equation}
In such decoupled joint modeling, the convolution weights utilized by each joint are independent and do not interfere with each other.
Thus, it can learn the distinct motion attributes of different joints and enhance the temporal semantic representations. 
The joint-decoupled modeling is a general temporal modeling strategy for long skeleton sequences.
In the following, we extend the JTM module by the dilated residual TCNs~\cite{farha2019ms} and the linear transformer~\cite{katharopoulos20a}. 

\myPara{Temporal modeling with dilated residual TCNs.}
Following the previous approaches~\cite{farha2019ms,ishikawa2021alleviating}, we employ dilated residual TCNs as $\mathcal {T}(\cdot)$ to capture the temporal features.
Specifically, the computation of TCNs can be formally described as follows:
\begin{equation}
 \mathcal {T}(J) = \sum\limits_{v=1}^V{\rm DConv}(W^J_v, J_v^{}) + J_v^{},
\label{eq6}
\end{equation} 
where ${\rm DConv}(W^J_v, J_v^{})$ denotes the dilated convolutional operation with weights $W^J_v$ and input $J_v^{}$. 
By stacking multiple layers of JTM, both the short-distance and long-distance temporal relationships between joints can be captured.

\input{figs/4-JTM}

\myPara{Temporal modeling with linear transformer.}
Unlike TCNs that capture features under a fixed temporal receptive field, the linear transformer~\cite{katharopoulos20a} can be applied as $\mathcal {T}(\cdot)$ to model the global temporal dependencies of joints.
Formally, the linear transformer layer can be computed as follows:
\begin{equation}
 {T}(J) = \sum\limits_{v=1}^V \phi(Q_v^{})(\phi(F_v^{})^T U_v^{}) + J_v^{},
\label{eq7}
\end{equation}
where $Q_v^{}$, $F_v^{}$, and $U_v^{}$ are transformed from $J_v^{}$ by the linear layers, and $\phi(\cdot)$ is the activation function.
Compared to the previous methods~\cite{chinayi_ASformer} that use sliding window attention, the linear transformer is more suitable for modeling long sequences due to its linear complexity.

% \myinputalgorithm{algorithm}
  
\subsection{Decoupled spatio-temporal Framework}
\label{sec:DSTF}

\myPara{Overall architecture.}
By combining the DSTI and JTM modules, we construct a powerful spatio-temporal framework, \ie \ourMthd, for skeleton-based action segmentation.
The full architecture of the proposed \ourMthd~is shown in~\figref{Framework}.
% 
% Algorithm \ref{JFormer} also summarizes the main implementation.
% 
Specifically, the spatial features generated by unified spatial modeling are divided into $M$ groups of spatial sub-features.
Then, $L_y$ layers of Joint-decoupled Temporal Modeling (JTM) are used to capture the rich temporal relationships among joints.
The first layer of JTM takes the first spatial sub-feature $\hat S_1$ as input, capturing the temporal dependencies of each joint.
In the next $L_y-1$ JTM layers, such as the $l$-th layer JTM, the input is the spatio-temporal features $R_{l-1}$, which are generated from the decoupled spatio-temporal interaction involving spatial sub-features of the $l$-th group and the temporal features from $(l-1)$-th layer of JTM.
Overall, the above process can be summarized as follows:
\begin{equation}
  \begin{split}
  H_{(l)} = \begin{cases}\mbox {$\mathcal T$}(\hat S_{1}), & \text { if } l=1, \\ 
    \mbox {$\mathcal T$}(R_{l-1}), & \text { if } 1< l \leq L_y, \\
    % \mbox {$\mathcal T$}(H_{(l-1)}) & \text { otherwise,}
    \end{cases}
\label{eq9}
\end{split}
\end{equation}
where $H_{(l)}$ is the output features of the $l$-th layer in \ourMthd.
Then,  we attain the spatio-temporal features $H_{(L_y)}$ from the last layer.
To predict the probability for the action category and action boundary of each frame, we apply two 1D convolutions over the spatio-temporal features $H_{(L_y)}$ as follows:
\begin{equation}
  \begin{aligned}
    Y_{c} = {\rm Softmax}({\rm Conv1D} (H_{(L_y)})),
  \end{aligned}
\label{eq10}
\end{equation}
\begin{equation}
  \begin{aligned}
    Y_{b} = {\rm Sigmoid}({\rm Conv1D} (H_{(L_y)})),
  \end{aligned}
\label{eq11}
\end{equation}
where $Y_{c} \in {R^{{C^{o}} \times T }}$ denotes the probabilities of the action category, $C^{o}$ is the total number of action categories, 
and $Y_{b} \in {R^{ {1} \times T }}$ represents the probability that a frame is at the edge of the action. 
Following ASRF~\cite{ishikawa2021alleviating,chinayi_ASformer}, to improve the performance of action segmentation, we further refine the initial predictions $Y_{c}$ and $Y_{b}$ via action segmentation branch~(ASB) and boundary regression branch~(BRB), respectively. 
In~\ourMthd, the ASB and BRB branches utilize two and three stages of temporal modeling stages~\cite{ishikawa2021alleviating,chinayi_ASformer}, respectively.

\input{tabs/II-Datasets}

\input{tabs/III-MCFS}

\myPara{Loss function.}
We use cross entropy for the action segmentation branch as the classification loss $\mathcal{L}_{\rm ce} $ following existing works~\cite{farha2019ms,ishikawa2021alleviating}.
To penalize over-segmentation errors, the Gaussian similarity-weighted TMSE loss $\mathcal{L}_{{\rm gs-tmse}}$~\cite{ishikawa2021alleviating} is used to smooth the transition of class probabilities between frames.
Overall, the loss function for predictions at each stage in the action segmentation branch is defined as follows:
\begin{equation}
  \begin{aligned}
    \mathcal{L}_{{\rm asb}} = \mathcal{L}_{{\rm ce}} + \mathcal{L}_{{\rm gs-tmse}}.
  \end{aligned}
  \label{eq12}
\end{equation}
For the boundary regression branch, we use a binary logistic regression loss $\mathcal{L}_{{\rm brb}}$ at each stage as follows:
\begin{equation}
\begin{aligned}
\mathcal{L}_{{\rm brb}} = \frac{1}{T} \sum_{t=1}^{T}(w_{p} P_{b}(t) \cdot \log Y_{b}(t)+\\
\left(1-P_{b}(t)\right) \cdot \log \left(1-Y_{b}(t)\right)),
\end{aligned}
\label{eq13}
\end{equation}
where $P_{b}(t)$ is the ground truth that only encodes the value of 1 at the action boundary.
$w_{p}$ is the weight of positive samples (\ie the reciprocal of the number of boundaries over the whole frames), which balances the weights of the number of boundary frames and other frames.
In summary, the action segmentation branch and boundary regression branch can be jointly trained with the loss function as follows:
\begin{equation}
  \begin{aligned}
    \mathcal{L} = \mathcal{L}_{{\rm asb}} + \gamma \mathcal{L}_{{\rm brb}},
  \end{aligned}
  \label{eq14}
\end{equation}
where $\gamma$ is the loss weight that is set to 0.1 by default.

\section{Experiments}
\subsection{Datasets}

We evaluate the proposed \ourMthd~for skeleton-based action segmentation on four datasets, including MCFS-22~\cite{liu2021temporal}, MCFS-130~\cite{liu2021temporal}, PKU-MMD~\cite{liu2017pku}, and LARa~\cite{niemann2020lara}. 
The statistics of these datasets are listed in~\tabref{datasets}.

\myPara{MCFS-22.}
The MCFS-22 is a high-quality action segmentation dataset containing 271 long sequences of the skeleton action.
Each long video consists of an average of 12 action segments, with an average duration of 212 seconds. 
The action segments are categorized into 22 classes.
Each skeleton graph contains $V = 25$ body joints extracted by the OpenPose toolbox~\cite{Cao_2017_CVPR}.

\myPara{MCFS-130.}
Compared to the MCFS-22, the MCFS-130 involves more fine-grained actions in both spatial and temporal dimensions and contains 130 action categories.
% 
% The actions of jumping and rotating perform different numbers of rotations. 
% (\eg ``Luzt2'' and ``Flip3'').

\myPara{PKU-MMD.}
The PKU-MMD is a large-scale dataset focusing on long continuous sequences for human action understanding.
It contains 42 action categories and 1009 long MoCap sequences recorded in three camera views.
MoCap sequences are recorded as the 3-axis locations of 25 body joints via the Kinect V2 sensor.
This dataset contains two benchmarks:
(1) X-sub (X-sub) where training data comes from 775 videos and testing data comes from the other 234 videos.
(2) X-view (X-view) where training data comes from the middle and right camera views and testing data comes from the left camera view.

\myPara{LARa.}
The LARa is a continuous action dataset where 14 subjects carry out typical warehousing activities.
It contains 377 long videos covering 8 action categories.
The actions were performed under 3 different real-world warehousing scenarios.
Following the settings of previous works~\cite{filtjens2022skeleton}, we down-sample the sampling frequency of action sequences from 200 fps to 50 fps.

\subsection{Evaluation Metrics}
Following previous works, we report three evaluation metrics, \ie frame-wise accuracy (Acc), segmental edit score, and segmental F1 score at the intersection over union thresholds 0.10, 0.25, and 0.5~(F1@10, F1@25, F1@50).
For MCFS-22 and MCFS-130 datasets, the 5-fold cross-validation is used for evaluation, 
and the averaged results are reported as shown in \tabref{MCFS}.
For PKU-MMD (X-sub), PKU-MMD (X-view), and LARa datasets, 
the single validation is used for evaluation, 
and the results are reported as shown in~\tabref{PKU_sub}, \tabref{PKU_view}, and \tabref{LARa}.

\subsection{Implementation Details}
All experiments are conducted on a single RTX 3090 GPU with the PyTorch~\cite{paszke2017automatic,paszke2019pytorch} deep learning framework.
By default, DSTF adopts the linear transformer~\cite{katharopoulos20a} for temporal modeling in JTM.
The hyper-parameters $M$ and $L_c$ defined in~\secref{sec:DSTF} are set to 10 and 10, respectively, and are discussed in ablation studies (\secref{sec:Ablation}).
We use an Adam optimizer for training on all datasets and keep all other hyper-parameters the same as~\cite{ishikawa2021alleviating}.
For MCFS-22 and MCFS-130, we train the models for 300 epochs with a learning rate of 0.0005, a batch size of 1, and the data pre-processing in~\cite{liu2021temporal}.
For the PKU-MMD and LARa datasets, we train the models for 150 epochs, where the learning rate is 0.001, the batch size is 8, and the data pre-processing follows~\cite{filtjens2022skeleton}.
In the ablation study, all experiments use the above settings for a fair comparison.

\input{tabs/IV-PKU_sub}
\input{tabs/V-PKU_view}

\subsection{Comparison with the State-of-the-Art}
We compare the proposed \ourMthd~with existing state-of-the-art methods, including temporal modeling methods~\cite{farha2019ms,Chen_2020_CVPR,wang2020boundary,ishikawa2021alleviating,chinayi_ASformer} and spatio-temporal modeling methods~\cite{filtjens2022skeleton,liu2022spatial,li2023involving}, on the MCFS-22, MCFS-130, PKU-MMD (X-sub), PKU-MMD (X-view), and LARa datasets, respectively.

\myPara{Comparison with Temporal Modeling Methods.}
We first compare \ourMthd~against temporal modeling methods, including TCN-based methods~(\eg ASRF~\cite{ishikawa2021alleviating}, BCN\cite{wang2020boundary}, and MS-TCN \cite{farha2019ms}) and Transformer-based methods~(\eg ASFormer~\cite{chinayi_ASformer}). 
Compared to the proposed \ourMthd, these methods cannot model spatial information of joints that is essential for skeleton-based action segmentation, thus limiting their performance. 
As shown in~\tabref{MCFS}, \ourMthd~consistently outperforms these methods. 
For example, on MCFS-130 in terms of F1@50, \ourMthd~that employs TCNs~\cite{farha2019ms} in temporal modeling outperforms the TCN-based ASRF by 7.3\%.  
\ourMthd~that employs the linear transformer~\cite{katharopoulos20a} in temporal modeling also outperforms the Transformer-based ASFormer by 7.5\%. 

\input{tabs/VI-LARa}

\input{figs/7-Flops-MCFS_130}

\myPara{Comparison with spatio-temporal Modeling Methods.}
Here, we compare our method with the approaches based on spatio-temporal modeling.
Even though these methods have implemented spatial modeling, \ourMthd~still outperforms previous competitive methods on MCFS-22 and MCFS-130, as shown in \tabref{MCFS}.
For example, on the challenging MCFS-130 dataset, \ourMthd~outperforms the previous SOTA method IDT-GCN by 5.1\% in terms of F1@10 with less computational complexity.
On MCFS-22, \ourMthd~also achieves a significant advantage in single-frame accuracy and segmental edit distance scores compared to IDT-GCN, while performing similar F1 scores with IDT-GCN.
This is because IDT-GCN employs an additional Temporal Segment Regression (TSR) branch that models action sequences for recognizing similar frames but requires more computational resources.
We also add the TSR branch to \ourMthd, \ie \ourMthd$^\ast$ in \tabref{MCFS}, which results in state-of-the-art performance on both MCFS-22 and MCFS-130.

\input{figs/6-Visualization-MCFS}

To further explore the effectiveness of our \ourMthd, we show the comparison on more datasets, including PKU-MMD and LARa. 
As shown in~\tabref{PKU_sub},~\tabref{PKU_view}, and~\tabref{LARa}, we can observe that \ourMthd~consistently achieves the best performance, especially on F1 scores. 
These results demonstrate that \ourMthd~predicts more precise and complete action segments across different scenes.

\myPara{Efficiency analysis.} 
To validate the efficiency, we compare the proposed \ourMthd~with existing methods in terms of the number of parameters and GFLOPs.
As shown in~\figref{Flops}, our method achieves state-of-the-art performance without sacrificing computational efficiency.
In particular, \ourMthd$^\dag$~is superior to the previous methods and is lightweight on FLOPs and the number of parameters (\eg 2.5 × less FLOPs than IDT-GCN and 8.0 × less FLOPs than MS-GCN).
\ourMthd~and \ourMthd$^*$~further achieve higher performances with less computation costs than the previous state-of-the-art methods.

\myPara{Qualitative results}.
We further provide qualitative analysis.
As shown in \figref{Visualization-MCFS}, the previous action segmentation methods fail to accurately locate the actions in the long sequence.
Especially, the fine-grained actions in MCFS-130 share similar appearance and motion patterns, leading some methods to produce incorrect predictions.
For example, as shown in \figref{Visualization-MCFS}(a), BCN and ETSN produce over-segmentation errors,~\ie segmenting a complete action clip into different categories.
ASRF also falsely detects the action boundaries, and IDT-GCN loses some action segments, as shown in \figref{Visualization-MCFS}(b).
MS-GCN also generates a lot of over-segmentation errors, as shown in \figref{Visualization-MCFS}(c) and (d).
These incorrect predictions limit the application of previous methods in practical applications.
In contrast, the proposed \ourMthd~effectively captures discriminating spatio-temporal features by decoupling cascaded interaction and joint-decoupled temporal modeling, thus relieving these incorrect predictions and achieving better segmentation results.
All the above competitive performances and good segmentation results verify the effectiveness of the \ourMthd.

\subsection{Analysis of Decoupled spatio-temporal Interaction}
\label{sec:exp-DSTI}

\myPara{Effects of spatial sub-features.} 
% We first visualize the divided spatial sub-features corresponding to the joint values on the human body.
The sub-features introduced in \secref{sec:STI} contain diverse spatial information and are used to interact with the temporal features from different layers. 
To analyze the diversity, we first visualize the activation values of different body joints. 
% Across different groups of sub-features, 
% we first visualize the activation values of different body joints. 
% the divided spatial sub-features corresponding to the joint values of the human body.
% 
As shown in~\figref{Visualization-Groups-sub-features}, different groups of sub-features focus on different joints.
For example, in action (a), the first two groups focus on the bottom \emph{foot}, and the last two groups focus on the top \emph{foot}. 
Meanwhile, \figref{Visualization-Groups-sub-features} also shows that the spatial sub-features can adapt to different actions. 
For instance, the same group pays more attention to the \emph{foot} and \emph{leg} for action (a) but focuses more on the \emph{foot} and \emph{hand} for action (b).
This phenomenon indicates that different groups of spatial sub-features can capture diverse and discriminative spatial features, effectively facilitating the spatio-temporal interaction.

% Meanwhile, 
% \figref{Visualization-Groups-sub-features} also shows that 
% spatial sub-features can learn the value of important joints related to actions.
% % 
% For instance, the spatial sub-features for action (a) pay more attention to the \emph{foot} and \emph{leg}, and the spatial sub-features for action (b) concern both \emph{foot} and \emph{hand}.
% % 
% This indicates that the spatial sub-features in DSTI can capture distinct and reasonable spatial features, contributing to the interaction with the temporal features.

\input{figs/Groups-sub-features}
\input{tabs/X-SSM-STI}

\myPara{Different types of spatio-temporal interaction.} 
To verify the effectiveness of the proposed DSTI module, we compare different interaction methods, including the choice of spatial features and the way of spatio-temporal interaction.
% 
% We use the same baseline model introduced in~\secref{sec:exp-DSTI}.
% 
As shown in~\tabref{SSM-STI}, 
1) Compared to the same spatial features interacting with temporal features from different layers, employing spatial sub-features significantly improves the performance. 
This phenomenon indicates that the temporal features from different layers require to interact with 
distinct spatial features, which can be provided by our spatial sub-features. 
2) Regarding the interaction scheme, cross-attention outperforms summation, demonstrating the advantage of adaptive interaction. 
3) Through combining the spatial sub-features and cross-attention,~\ie our proposed decoupled spatio-temporal interaction architecture, we can achieve the best performance. 

% As shown in~\tabref{SSM-STI}, 1) Compared to the baseline model, utilizing the same spatial features to interact with temporal features from different layers via summation~(model A in~\tabref{SSM-STI}) does not improve the performance.
% % 
% Differently, employing spatial sub-features~(model B in~\tabref{SSM-STI}) improves the performances significantly. 
% % 
% The improved performances indicate that distinct spatial features~(\ie our spatial sub-features) can help to interact with different temporal features. 
% % 
% 2) On the other hand, adopting the cross-attention scheme~(model C in~\tabref{SSM-STI}) also improves the performance, demonstrating the advantage of adaptive interaction.
% % 
% 3) Compared with models B and C, the model D in~\tabref{SSM-STI} interacts with spatial sub-features and temporal features by the cross-attention scheme and achieves 

% erior performances. 
% % 
% These results confirm that divided spatial features and adaptive interactions are equally important for segmenting actions in long skeleton sequences.

% To further validate the fine-grained modeling capability of models, we only evaluate the methods on fine-grained actions from MCFS-130 dataset.
% % 
% As shown in~\tabref{SSM-STI-Fine_grained}, model D still performs well compared to other spatio-temporal models.
% % 
% This indicates that our model has the ability to capture the discriminative spatio-temporal relations, leading to better recognition of fine-grained actions.

\input{figs/13-Visualization-Compared-Tsne}
% \myinputtab{XI-SSM-STI-Fine_grained}

\myPara{Qualitative comparison.}
Other than the above quantitative analysis, we provide qualitative results for a more comprehensive analysis.
Firstly, we utilize t-SNE~\cite{van2008visualizing} to show the embedding distribution of different layers generated by MS-GCN~\cite{filtjens2022skeleton} and our approach. 
As shown in~\figref{Visualization-Compared-Tsne}(a), we find that along with the increase of MS-GCN layers, the representations between different categories tend to be consistent, especially making action boundaries ambiguous.
This shows that as the number of spatio-temporal blocks increases, the problem of over-smoothing is more likely to occur in the methods based on cascaded interaction.
In contrast, through decoupling the cascaded interaction, our proposed \ourMthd~alleviates the problem of over-smoothing features and achieves more separable category boundaries.
Specifically, with the increase of network layers, the features from the same category are gradually pulled together, and the features from different categories are pushed away, indicating that our model could capture more discriminative spatio-temporal relations.
This phenomenon is also demonstrated in \secref{sec:Ablation}, where we show that using more layers of spatio-temporal interaction~(DSTI) leads to better performances. 
Meanwhile, the predictions corresponding to the features of ~\figref{Visualization-Compared-Tsne}(a) also verify that our model can segment action boundaries accurately, as shown in~\figref{Visualization-Compared-Tsne}(b).

% As shown in~\figref{Visualization-STI-Label}, the baseline, model A, model B, and model C either incorrectly detect some action boundaries or assign incorrect action categories. 
% % 
% In contrast, model D can effectively capture the spatio-temporal features of the joints for better representations of motion, reducing the above-mentioned incorrect predictions and achieving better segmentations.
% 
% As shown in~\figref{Visualization-STI-Tsne}, the t-SNE maps also verify that the feature embeddings learned by our model~(model D) have more separable category boundaries, thus helping identify the action boundaries. 

% \myinputfig{12-Visualization-STI-Label}
% \myinputfig{13-Visualization-STI-Tsne}

\subsection{Analysis of Joint-decoupled Temporal Modeling}
\label{sec:exp-JTM}

\myPara{Statistics on the motion speed of joints.}
To verify the motivation of the joint-decoupled temporal modeling, we count the movement speed of different joints in all sequences from MCFS-130.
% 
% For statistical methods, please refer to the supplementary material.
%
As shown in~\figref{Statistic-Joint-Speed}, different joints move at different speeds.
For example, compared to the joints near the spine, the joints in the shoulders and legs move faster and have a high variance in speed.
This phenomenon supports our motivation to model different joints separately.

\input{figs/8-Statistic-Joint-Speed}

\input{tabs/VII-JTM-JSTM}

\myPara{Comparison of joint-decoupled/shared weights.}
To compare the effect of independent or shared weights for modeling temporal information, we compare the previous temporal convolution networks~\cite{filtjens2022skeleton} that utilize dilated 2D convolution (kernel size is $1 \times 3$), denoted as 2D-Conv, to model all joints in shared weights way and our proposed JTM.
To focus on the temporal modeling capabilities of the model, we employ the baseline that only performs a single spatio-temporal interaction~\cite{liu2022spatial,li2023involving}. 
As shown in~\tabref{JTM-JSTM}, using TCN for timing modeling, the models with the JTM module achieve better performance than the models with 2D-Conv for all metrics in both MCFS-22 and MCFS-130.
For example, our JTM module can improve the 2D-Conv at 3.0\% on the F1@50 score on MCFS-130.

The above comparison is only conducted by utilizing TCN as temporal modeling.
To further explore the validity of JTM, we conduct a Joint-shared Weight Temporal Modeling~(JWTM) module that employs the dilated residual TCNs~\cite{farha2019ms} and the linear transformer~\cite{katharopoulos20a} and compare the JWTM with the proposed JTM.
% \lyh{To further explore the validity of JTM, we conduct a Joint-shared Weight Temporal Modeling~(JWTM) module that employs the dilated residual TCNs~\cite{farha2019ms} and the linear transformer~\cite{katharopoulos20a} and compare the JWTM with the proposed JTM.}
% 
% For the implementation of JWTM, please refer to the supplementary material. 
% 
For the implementation of JWTM, the input features $S^\prime \in \mathbb R^{C^\prime \times T \times V}$ are transformed into channel-level features $\hat S^{\prime} \in {\mathbb R^{C^\prime \times {T} }}$ via the $\mathcal {R}$ function.
Different from \eqcref{eq5}, 
the output features ${ \mathcal T({\hat S}^{\prime}) } \in {\mathbb R^{{C^t} \times T}}$ of the JWTM are represented as:
\begin{equation}
  {\mathcal T({\hat S}^{\prime})} = \sum\limits_{c=1}^{C^\prime} {(W^J_{c} \LARGE{\ast} \hat S^{\prime}_c)}.
\label{eq18}
\end{equation}
The temporal features generated by JWTM ignore the differences in the motion of different joints.
Unlike JTM which models each joint by independent weights, JWTM captures temporal dependencies of all joints using shared weights. 
As shown in~\tabref{JTM-JSTM}, the models with the JTM module output the model with the JWTM.
We also observe that the linear transformer using the JWTM results in a huge drop in performance on MCFS-22.
This indicates the importance of modeling the motion properties of each joint and the effectiveness of the JTM module in capturing the temporal dependencies of individual joints.

\input{figs/9-Visualization-JTM-JSTM-mean}
\input{tabs/IX-SSM-Transformation_functions}

\myPara{Qualitative analysis.}
To delve deeper into the difference between JTM and JWTM modules and how they affect the final performance, we visualize the predictions and temporal features extracted from the last layer when using JTM and JWTM, respectively. 
As shown in~\figref{Visualization-JTM-JSTM-mean}, regarding the predictions, we observe that our proposed JTM module identifies each action clip correctly. 
In contrast, the JWTM module fails to distinguish action boundaries and predicts the wrong action categories.
This is because the JWTM ignores differences in motion speed of different joints, thus weakening the temporal modeling semantics. 
This effect can also be seen from the temporal features, as shown in~\figref{Visualization-JTM-JSTM-mean}, where the learned features are too smooth and lack distinction. 
In contrast, the JTM enables the model to learn distinctive temporal dependencies from each joint, helping identify multiple action clips in long sequences.

\subsection{Ablation Analysis}
\label{sec:Ablation}
% In \secref{sec:exp-DSTI} and \secref{sec:exp-JTM}, we have ablated and analyzed different components of our proposed \ourMthd. 
% % 
% In this section, we further study the impact of the feature-transform functions as defined in~\secref{sec:STI} and 
% some hyper-parameters, including the number of JTM layers ($L_y$) defined in~\secref{sec:DSTF}.
% % 
% In addition, we also conduct ablation experiments about the number of DSTI layers, where only the first $L_c$ JTM layers, perform the interaction with spatial features and temporal features.

\myPara{Different feature transformation functions.} 
In \secref{sec:STI}, 
we propose the feature transformation function to align the dimensions between spatial and temporal features. 
Here, we explore the effects of different feature transformation functions. 
As shown in~\tabref{transformation_functions}, the models with Convolution and Avgpool outperform the models with Maxpool.
This is because the Maxpool is susceptible to outliers and may result in the loss of necessary information for identifying action boundaries.
In contrast, Avgpool retains overall information about the features, while Convolution processes the features adaptively using trainable weights.
Considering the performance of the model, we adopt the Convolution feature-transform function in \ourMthd.

\input{tabs/XII-JTM-layers}
\input{tabs/XIII-STI-layers}

\myPara{Impact of the number of JTM layers.}
The number of JTM, \ie $L_y$, defines the depth of the temporal modeling network for \ourMthd.
As shown in \tabref{JTM-layers}, increasing $L_y$ from 8 to 10 greatly boosts the performance, indicating that enough temporal modeling layers are essential.
Adding $L_y$ from 10 to 11 or 12 still improves the performance, but the improvements are less significant than increasing from 8 to 10. 
% 
% This might be an over-fitting problem caused by increasing stacked temporal modeling layers.

\myPara{Effect of the number of DSTI layers.} 
We also conduct ablation experiments about the number of DSTI layers, where only the first $L_c$ JTM layers, perform the interaction with spatial features and temporal features. 
As shown in \tabref{STI-layers}, the performance is significantly improved when increasing the $L_c$ from 1 to 9.  
This is mainly because the proposed decoupled interaction will not cause the problem of over-smoothing, thus making the spatio-temporal features more discriminative. 
% 
% Moreover, it confirms that our proposed decoupled spatio-temporal interaction patterns can effectively avoid the smoothing problem caused by stacking network layers.
%
To better understand the behavior of the decoupled spatio-temporal interaction layers, we also visualize the cross-attention maps from different layers of DSTI.
As shown in \figref{STI-CrossAttention-Map}, the attention maps vary across various layers. 
This demonstrates that our adaptive interaction module can learn the different interaction patterns at various layers, which is beneficial for generating distinctive spatio-temporal features. 

\input{figs/14-STI-CrossAttention-Map}

% \myinputtab{XI-SSM-STI-Fine_grained}

\section{Conclusion}

In this work, we present a Decoupled Spatio-Temporal Framework (\ourMthd) for skeleton-based action segmentation.
Unlike previous spatio-temporal models, \ourMthd~adopts decoupled spatio-temporal interaction and joint-decoupled temporal modeling.
The decoupled spatio-temporal interaction adopts a single layer of spatial modeling to extract spatial features, which are divided into different groups of spatial sub-features and then interact with temporal features from different layers, avoiding the over-smoothing caused by cascaded interaction. 
The joint-decoupled temporal modeling decouples the modeling of each joint, enabling the model to capture the distinct motion properties of different joints. 
% To avoid stacking multiple cascade interactions, the DSTI adopts single-layer unified spatial modeling and divides spatial features 
% into multiple spatial sub-features that contain different spatial semantics.
% 
% Then, for sufficient interaction with spatio-temporal information, the spatial sub-features adaptively interact with temporal features at different JTM layers via the cross-attention scheme.
% 
% To capture the distinct motion properties of different joints, 
% the JTM employs independent weights to model the temporal dependencies of each joint separately.
% 
On four large-scale benchmarks of different scenes, the proposed \ourMthd~achieves competitive performance with less computational costs than existing methods. 
% 
% Moreover, its FLOPs and number of parameters are less than other recent models, which is suitable for low computational cost devices.
% 
Due to the effectiveness and efficiency, the proposed \ourMthd~could serve as a new baseline for spatio-temporal modeling and have a large potential for developing more complex models.

\myPara{Limitation and Future work.}
Compared to previous methods, our proposed \ourMthd~has significantly improved accuracy in recognizing fine-grained actions, but there are still rooms to improve the performance.
In future, we will try fine-grained modeling in the spatial and temporal dimensions to further improve the performance.
For example, we can design new spatial topology to model complex spatial relationships of joints effectively and employ multi-stream fusion strategies~\cite{shi2020skeleton}.
Besides, since long action sequences are more accessible than edited action clips, we can also explore self-supervised long-sequence action modeling to get better pre-trained models.

\bibliographystyle{IEEEtran}
\bibliography{reference}

\vfill

\clearpage
\end{document}

%% file: figs/Introduction.tex
\begin{figure}[t]
  \centering
    \begin{overpic}[width=0.45\textwidth]{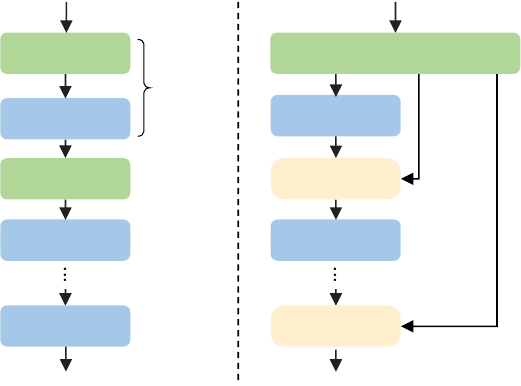}
    % (a) Existing Methods
    \put(14,71){\footnotesize{Skeleton Data}}
    \put(29,57.5){\footnotesize{\color{red} Cascaded}}
    \put(28.5,54){\footnotesize{Interaction}}
    \put(7,62.3){Spatial}
    \put(4.5,50){Temporal}
    \put(7,38.5){Spatial}
    \put(4.5,26.8){Temporal}
    \put(4.5,10){Temporal}
    \put(2,-2){\footnotesize{(a) Existing Methods}}
    % (b) DSTF
    \put(77.5,71){\footnotesize{Skeleton Data}}
    \put(63,62.3){Unified Spatial}
    \put(56.5,50.3){Temporal}
    \put(57,40.3){\footnotesize{\color{red} Decoupled}}
    \put(57,37){\footnotesize{Interaction}}
    \put(56.5,26.8){Temporal}
    \put(57,12){\footnotesize{\color{red} Decoupled}}
    \put(57,8){\footnotesize{Interaction}}
    \put(59,-2){\footnotesize{(b) \ourMthd~(Ours)}}
    \end{overpic}
   \vspace{1pt}
  \caption{Comparison of existing and our proposed methods for spatio-temporal modeling in the skeleton-based action segmentation.
  (a) Existing methods employ coupled spatial and temporal modeling for \textbf{cascaded} interaction.
  (b) In contrast, the proposed \ourMthd~\textbf{decouples} the cascaded spatial-temporal interaction by adopting unified spatial modeling to extract different groups of spatial sub-features, which adaptively interact with temporal features from different layers, respectively.
  % 
  % This decoupled spatio-temporal modeling avoids the effect of over-smoothing due to the coupled spatio-temporal model while achieving sufficient spatio-temporal interaction. 
  }
  \label{Introduction}
\end{figure}

%% file: figs/skeleton-Introduction.tex
\begin{figure}[t]
  \centering
    \begin{overpic}[width=0.40\textwidth]{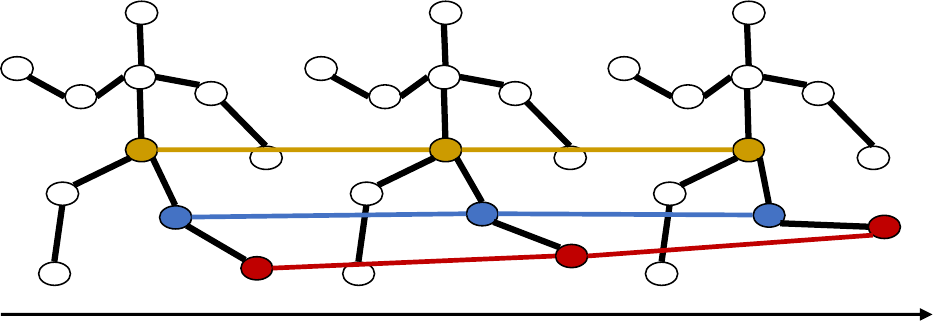}
    \put(43,-2){\footnotesize {Motion Flow}}
    \end{overpic}
   % \vspace{.5cm}
  \caption{Illustration of motion of body joints.
  The various joints exhibit distinct motion speeds.
  }
  \label{skeleton-Introduction}
\end{figure}

%% file: figs/Framework.tex
\begin{figure*}[t]
  \centering
    \begin{overpic}[width=0.97\textwidth]{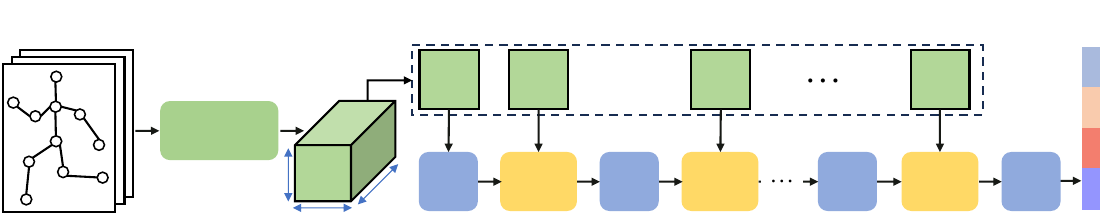} 
    \put(0,-3){\footnotesize {Skeleton Sequence}}
    \put(55,16){\footnotesize {$M$-group of Spatial Sub-features}}
    \put(39,16){\scriptsize {$V \times T$}}
    \put(14.7,8.5){\footnotesize \textbf{Unified Spatial}}
    \put(17,6){\footnotesize \textbf{Modeling}}
    \put(29.5,15){\footnotesize \textbf{Decouple}}
    \put(29,13){\footnotesize \textbf{Transform}}
    
    \put(26,-3){\footnotesize {Spatial Features}}
    \put(28.5,-0.5){\scriptsize {$V$}}
    \put(24.5,3.5){\scriptsize {$T$}}
    \put(34.2,1.8){\scriptsize {$C^s$}}
    \put(39,7.5){\scriptsize $\hat S_1$}
    \put(38.8,2.6){\small \textbf{JTM}}
    \put(47,7.5){\scriptsize $\hat S_2$}
    % \put(47.3,2.5){\small \textbf{STI}}
    \put(46.8,2.5){\small \textbf{DSTI}}
    \put(55.3,2.5){\small \textbf{JTM}}
    \put(63.5,7.5){\scriptsize $\hat S_3$}
    % \put(63.7,2.5){\small \textbf{STI}}
    \put(63.2,2.5){\small \textbf{DSTI}}
    \put(75,2.5){\small \textbf{JTM}}
    \put(82.5,7.5){\scriptsize $\hat S_M$}
    % \put(83.8,2.5){\small \textbf{STI}}
    \put(83.2,2.5){\small \textbf{DSTI}}
    \put(91.8,2.5){\small \textbf{JTM}}
    
    \put(94,-3){\footnotesize Predictions}
    
    \end{overpic}
   \vspace{.5cm}
  \caption{Framework of the proposed end-to-end Decoupled Spatio-Temporal Framework (\ourMthd).
  % 
  % The input to \ourMthd~is skeleton sequence ${\mathcal V}$.
  % 
  We first employ unified spatial modeling networks (introduced in \secref{sec:SSM}) to capture spatial features $S \in {\mathbb R^{C^s \times T \times V }}$ of the joints, where $C^s$, $T$, and $V$ denote the numbers of output channels, frames, and joints, respectively.
  Then, the obtained spatial features $S$ are divided into a group of spatial sub-features $\hat S_i \in {\mathbb R^{V \times T}}$ that
  could learn different spatial semantics. 
  Moreover, the JTM (introduced in \secref{sec:JTM}) module is utilized to model the temporal dependencies of each joint individually, achieving discriminative temporal features. 
  In the DSTI (introduced in \secref{sec:STI}) block, spatial sub-features adaptively interact with temporal features, capturing complex spatio-temporal representations.
  }
  \label{Framework}
\end{figure*}
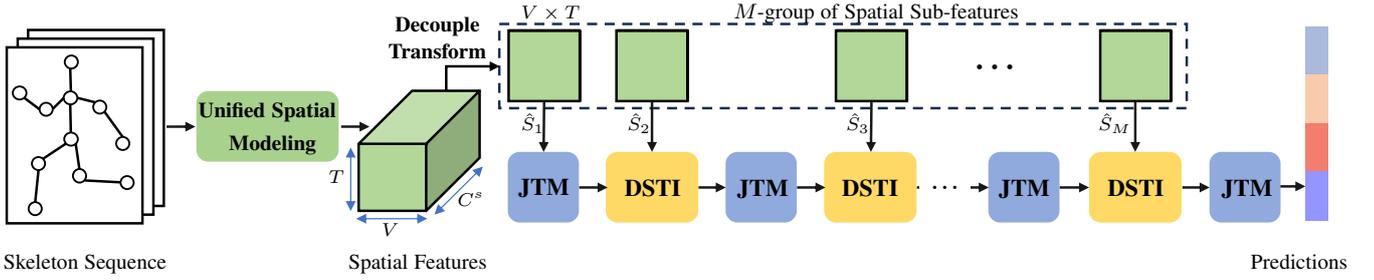

%% file: tabs/I-Symbols.tex
\begin{table}[t]
  \centering
  \setlength{\abovecaptionskip}{0pt}
  \caption{The definition and description of symbols. 
 }
\setlength{\tabcolsep}{0.5mm}{
  \begin{tabular}{lcl}
      \toprule
      Symbols & Dimensions/Type & Meaning \\
      \midrule
      $V$ & scalar & Number of joints.\\
      $T$ & scalar & Number of frames.\\
      $C^{},C^s,C^t$ & scalar & Number of Channel.\\
      \midrule
      \multicolumn{3}{l}{Multi-scale Spatial Modeling~(\textbf{MSM}) in \secref{sec:SSM}}  \\
      \midrule
      $X$ & $C \times T \times V$ & Joint features. \\
      $\hat A$ & $V \times V$ & Normalized adjacency matrix.\\
      $A^{(k)}$ & $V \times V$ & $k$-adjacency matrix. \\
      $B^{(k)}$ & $V \times V$ & Learnable matrix. \\
      \midrule
      \multicolumn{3}{l}{Decoupled Spatio-Temporal Interaction~(\textbf{DSTI}) in \secref{sec:STI}}  \\
      \midrule
      $S$ & $C^{s} \times T \times V$ & Spatial features output by spatial modeling.\\
      $S_i$ & $\frac{C^{s}}{M} \times T \times V$ & Spatial sub-features.\\
      $\hat{S}_i$ & $V\times T$ & Joint-level features transform from $S_i$. \\
      $R_{}$ & ${C^t}\times T$ & Spatio-temporal features output by STI.\\
      $M$& scalar&  Number of groups of the spatial sub-features. \\
      \midrule
      \multicolumn{3}{l}{Joint-wise Temporal Modeling~(\textbf{JTM}) in \secref{sec:JTM}} \\
      \midrule
      $J$ & $V \times T$& Input features for JTM.\\
      $\mathcal T(J)$ & $C^t \times T$& Temporal features output by JTM.\\
      % {$\ast$} & operation&convolution operation. 
      % \\
      % \midrule
      % \multicolumn{3}{l}{Decoupled Spatio-Temporal Framework~(\textbf{\ourMthd}) in \secref{sec:DSTF}}  \\
      % \midrule
      % $H_{(l)}$& ${C^t}\times T$ & Spatio-temporal features output by \ourMthd~layer.\\
      \bottomrule
  \end{tabular}}
  \label{Symbols}
\end{table} 

%% file: figs/Transform.tex
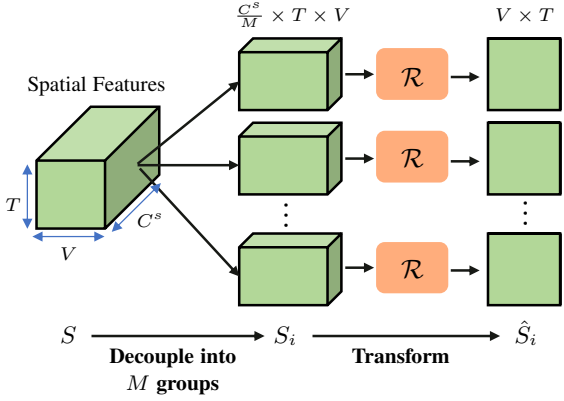
\begin{figure}[t]
  \centering
    \begin{overpic}[width=0.40\textwidth]{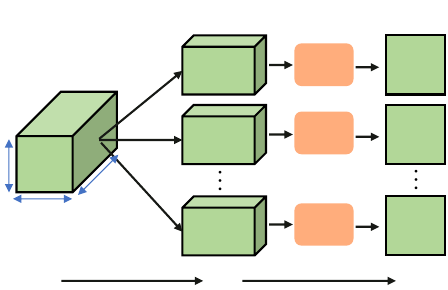}
        \put(2,47){\footnotesize {Spatial Features}}
        \put(8,16){\scriptsize {$V$}}
        \put(-2,25){\scriptsize {$T$}}
        \put(22,22){\scriptsize {$C^s$}}
        \put(39.8,60){\scriptsize {$\frac{C^{s}}{M} \times T  \times V$}}
        \put(87.5,60){\scriptsize {$V \times T$}}
        \put(70.3,48.0){$\mathcal {R}$}
        \put(70.3,33.0){$\mathcal {R}$}
        \put(70.3,12.7){$\mathcal {R}$}
        \put(8,0.7){\small {$S$}}
        \put(47,0.7){\small {$S_i$}}
        \put(91,0.7){\small {$\hat S_i$}}
        \put(17,-3){\footnotesize \textbf{Decouple into}}
        \put(20,-8){\footnotesize \textbf{$M$ groups}}
        \put(61.3,-3){\footnotesize \textbf{Transform}}
    \end{overpic}
   \vspace{.5cm}
  \caption{Illustration of decoupling and transforming the spatial sub-features.
  The $\mathcal {R}$ function, as defined in~\secref{sec:STI}, transforms the spatial sub-features to align with the number of dimensions of temporal features.
  Then, in~\secref{sec:STI}, the transformed spatial sub-features interact with temporal features from different layers.
  % 
  % The features $S$, $S_i$, and $\hat S_i$ are defined in \tabref{Symbols}.
  % 
  % \gsh{If you have space, I suggest you explain the meaning of the figure in the caption. Or remove unnecessary details. Some reviewers would just read the figure and caption to get your idea.}
  }
  \label{Transform}
\end{figure}

%% file: figs/4-JTM.tex
% \begin{figure}[t]
%   \centering
%   \begin{overpic}[width=0.40\textwidth]{images/4-JTM}
%     \put(2,40){\scriptsize {Joint 1}}
%     \put(12,35){\scriptsize {$J_1$}}
%     \put(2,29){\scriptsize {Joint 2}}
%     \put(12,24){\scriptsize {$J_2$}}
%     \put(2,18){\scriptsize {Joint 3}}
%     \put(12,13){\scriptsize {$J_3$}}
%     \put(2,7){\scriptsize {Joint 4}}
%     \put(12,2){\scriptsize {$J_4$}}
%     \put(29.2,34.5){\LARGE $\ast$}
%     \put(29.2,23){\LARGE $\ast$}
%     \put(29,12.5){\LARGE $\ast$}
%     \put(29,1.5){\LARGE $\ast$}
%     % 
%     \put(42,35){\scriptsize {$W^J_1$}}
%     \put(42,23){\scriptsize {$W^J_2$}}
%     \put(42,13){\scriptsize {$W^J_3$}}
%     \put(42,2){\scriptsize {$W^J_4$}}
%     % 
%     \put(80,36){\scriptsize {$\mathcal T(J)_1$}}
%     \put(80,28){\scriptsize {$\mathcal T(J)_2$}}
%     \put(80,20){\scriptsize {$\mathcal T(J)_3$}}
%     \put(80,12){\scriptsize {$\mathcal T(J)_4$}}
%     \put(80,3){\scriptsize {$\mathcal T(J)_5$}}
%     % 
%     \put(69,48){\scriptsize {joint-wise weights}}
%     \put(69,43){\scriptsize {features}}
%   \end{overpic}
%    % \vspace{-.2cm}
%   \caption{The process of joint-wise temporal modeling. 
%   % 
%   {$\ast$} denotes the convolution operation.
%   }
%   \label{JTM}
% \end{figure}

\begin{figure}[t]
  \centering
  \begin{overpic}[width=0.48\textwidth]{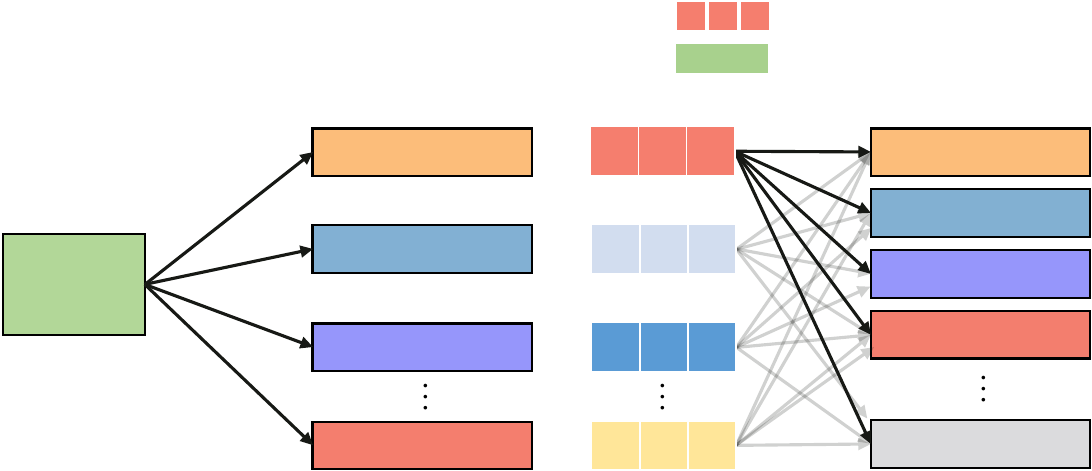}
    \put(-1,23){\scriptsize {Input Features}}
    \put(0,-5){\small {$J \in {\mathbb R^{V \times T}}$}}
    % \put(50,-5){\small {$W^J_v \in {\mathbb R^{{C^t} \times {C^f}}}$}}
    \put(77,-5){\small {$\mathcal T(J) \in {\mathbb R^{{C^t} \times T}}$}}
    \put(16,21){\rotatebox{40} {\scriptsize {\textbf{Decouple}}}}
    \put(29,32.2){\scriptsize {Joint 1~($1 \times T$)}}
    % \put(37,26){\scriptsize {$J_1$}}
    \put(29,23.2){\scriptsize {Joint 2}}
    % \put(37,17){\scriptsize {$J_2$}}
    \put(29,14.4){\scriptsize {Joint 3}}
    % \put(37,9){\scriptsize {$J_3$}}
    \put(29,5.3){\scriptsize {Joint $V$}}
    % \put(37,2){\scriptsize {$J_4$}}
    % 
    \put(50.6,27.8){$\ast$}
    \put(50.6,19){$\ast$}
    \put(50.6,10.3){$\ast$}
    \put(50.6,01.5){$\ast$}
    \put(54,32.7){\scriptsize {$W^J_1$}}
    \put(54,23.7){\scriptsize {$W^J_2$}}
    \put(54,15){\scriptsize {$W^J_3$}}
    \put(54,5.6){\scriptsize {$W^J_V$}}
    % 
    % \put(84,26){\scriptsize {$\mathcal T(J)_1$}}
    % \put(84,20){\scriptsize {$\mathcal T(J)_2$}}
    % \put(84,14){\scriptsize {$\mathcal T(J)_3$}}
    % \put(84,8){\scriptsize {$\mathcal T(J)_4$}}
    % \put(84,2){\scriptsize {$\mathcal T(J)_5$}}
    % 
    \put(72,41){\scriptsize {joint-decoupled weights}}
    \put(72,37){\scriptsize {features}}
  \end{overpic}
   \vspace{.5cm}
  \caption{The process of Joint-decoupled Temporal Modeling (\textbf{JTM}), where $V$, $T$, and $C^t$ denote the numbers of input channels, frames, and output channels, respectively.
  {$\ast$} represents the convolution operation.
  }
  \label{JTM}
\end{figure}
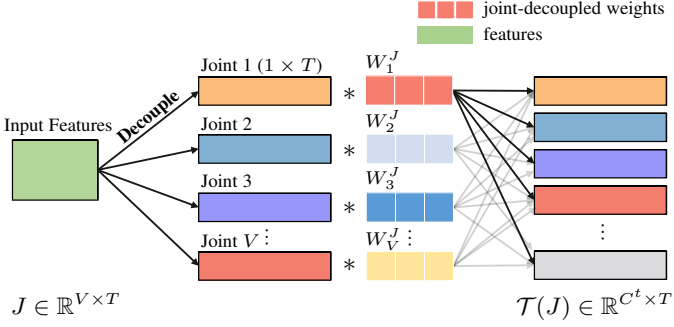

%% file: tabs/II-Datasets.tex
\begin{table}[htp!]
  \centering
  \setlength{\abovecaptionskip}{0pt}
  \caption{The statistics of action segmentation datasets. 
  Seq\# denotes the number of frames in an untrimmed long video. }
\setlength{\tabcolsep}{3.0mm}{
  \begin{tabular}{lccccc}
      \toprule
      {Dataset} & videos\# & class\# & clip\#& Seq\#& FPS\\
      \midrule
      {MCFS-22}& 271& 22& 12&$ \sim $6000& 30\\
      {MCFS-130}& 271& 130& 12&$ \sim $6000& 30\\
      {\makecell[c]{PKU-MMD}} & 1009& 42& 7&$ \sim $1000& 30\\
      {LARa} & 377& 8&22 &$ \sim $6000& 50\\
      \bottomrule
  \end{tabular}}
  \label{datasets}
\end{table} 

%% file: tabs/III-MCFS.tex
\begin{table*}[!t]
  \centering
  \setlength{\abovecaptionskip}{0pt}
  \caption{Comparison with the state-of-the-art on 
  MCFS-22~\cite{liu2021temporal} and MCFS-130 datasets~\cite{liu2021temporal}. 
  $^\dag$ represents that the model uses TCNs instead of the linear transformer 
  in the joint-wise temporal modeling~(JTM). 
  $^\ast$ indicates the model has an additional TSR~\cite{li2023involving} branch.
  }
  \setlength{\tabcolsep}{1.5mm}{
  \begin{tabular}{llcccccccccccc}
  \toprule
  & & & & \multicolumn{5}{c}{MCFS-130}&\multicolumn{5}{c}{MCFS-22}\\
  \cmidrule(lr){5-9}\cmidrule(lr){10-14}
  & Methods &FLOPs~$\downarrow$ &Param.~$\downarrow$ & F1$@10$~$\uparrow$ & F1$@25$~$\uparrow$ & F1$@50$~$\uparrow$ &Edit~$\uparrow$ &Acc~$\uparrow$ & F1$@10$~$\uparrow$ & F1$@25$~$\uparrow$ & F1$@50$~$\uparrow$ & Edit~$\uparrow$ &Acc~$\uparrow$\\
  \midrule
  \multirow{7}{*}{\rotatebox{-90}{Temporal}}
  &MS-TCN \cite{farha2019ms}& 4.04 G & 0.67 M & 56.4 & 52.2 & 42.5 & 54.5 & 65.7& 74.3 & 69.7 & 59.5 & 74.2 & 75.6  \\
  &SSTDA\cite{Chen_2020_CVPR} &- & -& 63.8 & 60.1 & 49.8 & 63.5 & 65.4& 76.7 & 72.2 & 61.2 & 77.5 & 75.7\\
  &ETSPNet\cite{li2021efficient}&4.41 G & 0.74 M & 60.9 & 56.6 & 47.5 & 59.8 & 65.6 & 74.3 & 69.9 & 59.4 & 73.8 & 76.6 \\
  &ETSN\cite{li2021efficient}&4.41 G&0.74 M& 64.5 & 61.0 & 52.3 & 64.6 & 64.6 & 81.4 & 77.6 & 66.8 & 79.8 & 77.0  \\
  &BCN\cite{wang2020boundary} &8.95 G&1.49 M& 60.4 & 55.9 & 46.4 & 58.8 & 68.1  &83.4&79.1&67.4&83.7&78.5\\
  &ASRF \cite{ishikawa2021alleviating}&7.04 G&1.17 M& 66.7 & 62.3 & 51.9 & 65.6 & 65.6 & 83.3 & 80.1 & 69.2 & 77.3 & 75.5 \\
  &ASFormer\cite{chinayi_ASformer}& 6.05 G & 1.06 M& 68.3 & 64.0 & 55.1 & 69.1 & 67.5  & 82.8 & 77.9 & 66.9 & 82.3 & 78.7 \\
  \midrule
  \multirow{8}{*}[1.0ex]{\rotatebox{-90}{Spatio-temporal}}
  &MS-GCN \cite{filtjens2022skeleton}& 39.97 G&1.10 M& 52.4 & 48.8 & 39.1 & 52.6 &  64.9&75.7 & 70.5 & 57.9 & 72.6 & 75.5 \\
  &SFA+MS-TCN \cite{liu2022spatial}& -& -   & -    & - & - & -    & -& 81.3 & 77.4 & 67.0 & 80.0 & 80.7 \\
  &SFA+ETSPNet \cite{liu2022spatial}   & -& -& -    & - & - & -    & -   & 82.1 & 78.3 & 68.6 & 80.8 & 81.4  \\
  &ID-GCN+ASRF  \cite{li2023involving}  & 8.24 G  & 1.27 M& 68.7 & 65.6 & 56.9 & 68.2 & 67.1  & 86.4 & 83.4 & 73.0 & 81.6 & 78.1 \\
  &IDT-GCN$^\ast$  \cite{li2023involving}& 12.81 G & 2.03 M& 70.7 & 67.3 & 58.6 & 70.2 & 68.6  &88.0 & 84.9 & 74.9 & 84.5 & 79.9 \\
  \cmidrule{2-14}
  &\ourMthd$^\dag$~(Ours) & 4.91 G & 0.76 M & 74.0 & 70.7 & 61.8 & 73.8 & 70.5& 86.6  & 83.5 & 73.2 & 82.3 & 78.7  \\
  &\ourMthd~(Ours) & 6.89 G & 1.10 M& 75.8 & 72.2 & 63.0 & 75.8 & 71.4  & 87.4  & 84.5 & 75.0 & \textbf{85.2} & 80.4 \\
  &\ourMthd$^\ast$ (Ours)  & 12.32 G & 2.00 M& \textbf{79.0} & \textbf{75.4} & \textbf{66.0} & \textbf{78.4} & \textbf{73.1}  &  \textbf{88.1}  &  \textbf{85.4} & \textbf{76.2} & 84.9 & \textbf{80.5} \\
  \toprule
\end{tabular}}
\label{MCFS}
\end{table*}

%% file: tabs/IV-PKU_sub.tex
\begin{table}[!t]
  \centering
  \setlength{\abovecaptionskip}{0pt}
  \caption{Comparison with the state-of-the-art on the PKU-MMD dataset 
  using the benchmark of x-sub. 
  }
\setlength{\tabcolsep}{1.5mm}{
  \begin{tabular}{lcccccc}
  \toprule 
  Methods & F1$@10$~$\uparrow$ & F1$@25$~$\uparrow$ & F1$@50$~$\uparrow$ & Edit~$\uparrow$ & Acc~$\uparrow$ \\
  \midrule
  Online-LSTM \cite{carrara2019lstm}&    -   &   -    & 23.3 & - & 57.7 \\
  TCN \cite{lea2017temporal}&    -   &   -    & 13.8 & - & 61.9 \\
  ST-GCN \cite{yan2018spatial}     &    -   &    -   & 15.5 & - & 64.9 \\
  MS-TCN \cite{farha2019ms}   &    -   &   -    & 46.3 & - & 65.5 \\
  MS-GCN \cite{filtjens2022skeleton}     &    -   &   -    & 51.6 & - & 68.5 \\
  CTC   \cite{xu2023efficient}& 69.9 & 66.4 & 53.8 & - & 69.2 \\
  \midrule
  \ourMthd$^\dag$~(Ours)& 71.7 & 68.0& 55.5 & 66.3 & 67.6 \\
  \ourMthd~(Ours)& \textbf{74.5} & \textbf{71.0} & \textbf{58.7} & \textbf{69.3} & \textbf{70.3} \\
  \bottomrule 
\end{tabular}}
\label{PKU_sub}
\end{table}

%% file: tabs/V-PKU_view.tex
\begin{table}[!t]
  \centering
  \setlength{\abovecaptionskip}{0pt}
  \caption{Comparison with the state-of-the-art on the PKU-MMD dataset using the benchmark of x-view. 
  ($^\ddag$ indicates our implemented results).
  }
\setlength{\tabcolsep}{1.8mm}{
  \begin{tabular}{lcccccc}
  \toprule 
  Methods & F1$@10$~$\uparrow$ & F1$@25$~$\uparrow$ & F1$@50$~$\uparrow$ & Edit~$\uparrow$ & Acc~$\uparrow$ \\
  \midrule
  MS-TCN \cite{farha2019ms}$^\ddag$& 58.6 & 53.6 & 39.4 & 56.6 & 58.2 \\
  ETSN \cite{li2021efficient}$^\ddag$& 62.4& 57.9 & 44.3 & 57.6 & 60.7 \\
  ASRF \cite{ishikawa2021alleviating}$^\ddag$& 62.5 & 58.0 & 46.1 & 59.3 & 60.4 \\
  MS-GCN \cite{filtjens2022skeleton}$^\ddag$& 61.3 & 56.7 & 44.1 & 58.1 & 65.3\\
  \midrule
  \ourMthd$^\dag$~(Ours)& 63.2 & 59.2& 47.6& 58.2& 62.4\\
  \ourMthd~(Ours) & \textbf{69.3} & \textbf{65.6} & \textbf{52.0} & \textbf{64.7} & \textbf{67.3} \\
  \bottomrule 
\end{tabular}}
\label{PKU_view}
\end{table}

%% file: tabs/VI-LARa.tex
\begin{table}[!t]
  \centering
  \setlength{\abovecaptionskip}{0pt}
  \caption{Comparison with the state-of-the-art on the LARa dataset. 
  }
\setlength{\tabcolsep}{1.8mm}{  
\begin{tabular}{lcccccc}
  \toprule 
  Methods & F1$@10$~$\uparrow$ & F1$@25$~$\uparrow$ & F1$@50$~$\uparrow$ & Edit~$\uparrow$ & Acc~$\uparrow$ \\
  \midrule
  Bi-LSTM  \cite{graves2005framewise}& -& - & 32.3& - & 63.9 \\
  TCN \cite{lea2017temporal} & -& - & 20.0& - & 61.5 \\
  ST-GCN \cite{yan2018spatial} & -& - & 25.8& - & 67.9 \\
  MS-TCN \cite{farha2019ms} & -& - & 39.6& - & 65.8 \\
  MS-TCN++ \cite{li2020ms} & -& - & 40.1& 52.3 & 63.4 \\
  BCN \cite{wang2020boundary} & -& - & 48.2& 57.3 & 65.4 \\
  ASRF \cite{ishikawa2021alleviating} & -& - & 50.2& 60.3 & 69.5 \\
  MS-GCN\cite{filtjens2022skeleton} & - & -& 43.6 & - & 65.6 \\
  STGA-Net\cite{tian2023stga} & - & -& 53.3 & 65.4 & 70.4 \\
  \midrule
  \ourMthd$^\dag$~(Ours)& 69.7 & 66.7 & 55.8 & 63.7 & 72.6 \\
  \ourMthd~(Ours) & \textbf{70.3} & \textbf{68.0} & \textbf{57.7} & \textbf{64.2} &  \textbf{75.1} \\
  \bottomrule 
\end{tabular}}
\label{LARa}
\end{table}

%% file: figs/7-Flops-MCFS_130.tex
\begin{figure}[!t]
  \centering
    \begin{overpic}[width=0.45\textwidth]{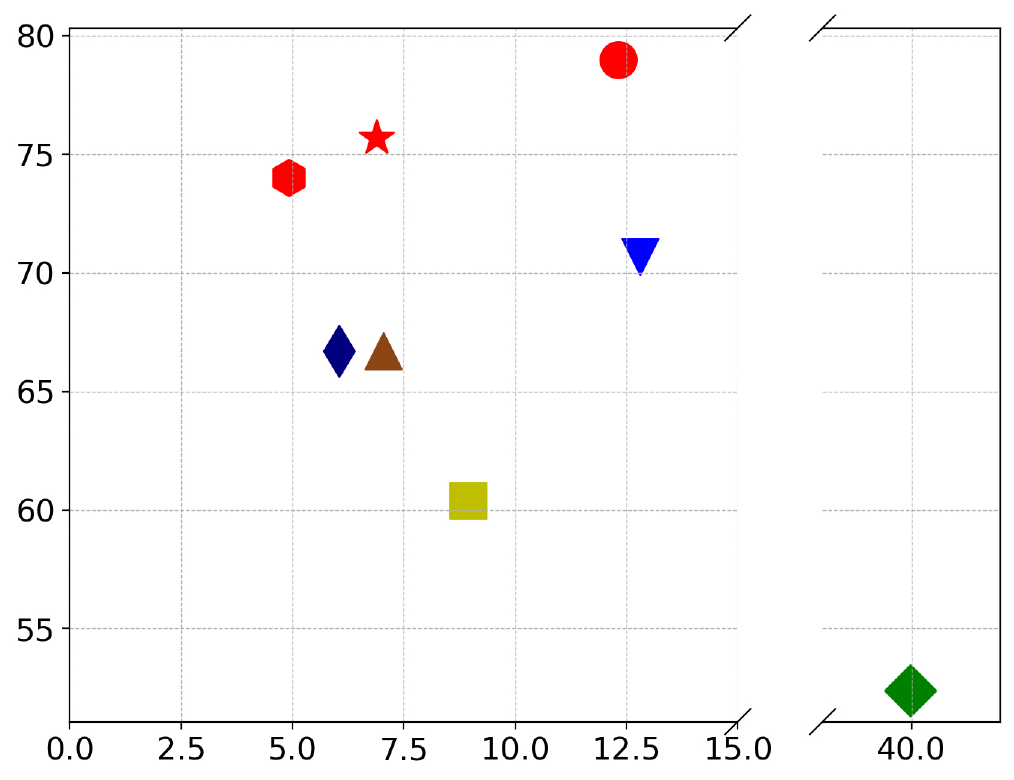}
    \put(30,67){\color{red}{\footnotesize {\ourMthd~(1.10M)}}}
    \put(8,62){\color{red}{\footnotesize {\ourMthd$^\dag$~(0.76M)}}}
    \put(64,69.5){\color{red}{\footnotesize {\ourMthd$^\ast$~(2.00M)}}}
    \put(64,54){\footnotesize {IDT-GCN (2.03M)}}
    \put(38,45){\footnotesize {ASRF (1.17M)}}
    \put(26,36){\footnotesize {ASFormer (1.06M)}}
    \put(49,26.5){\footnotesize {BCN (1.49M)}}
    \put(68,12){\footnotesize {MS-GCN (1.10M)}}
    \put(30,-2){ \footnotesize{GFLOPs/action sequence}}
    \put(-3,30){\rotatebox{90} {\footnotesize{F1@10 score}}}
    \end{overpic}
   % \vspace{.8cm}
  \caption{The F1@10 score and GFLOPs of different methods on the MCFS-130 dataset. 
  The values in the brackets indicate the number of parameters.
  The proposed \ourMthd~outperforms previous methods in terms of segmentation quality, parameter amounts, and GFLOPs.
  }
  \label{Flops}
\end{figure}
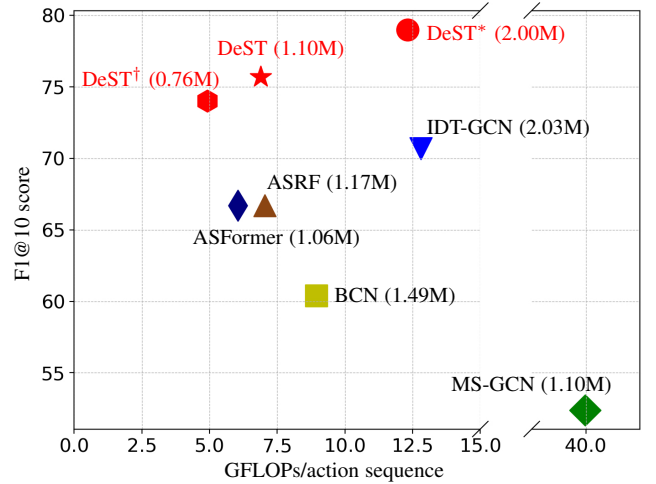

%% file: figs/6-Visualization-MCFS.tex
\begin{figure*}[t]
    \setlength{\abovecaptionskip}{0pt}
    \begin{minipage}[t]{0.49\linewidth}
        \flushright
        \begin{overpic}[width=0.8\textwidth]{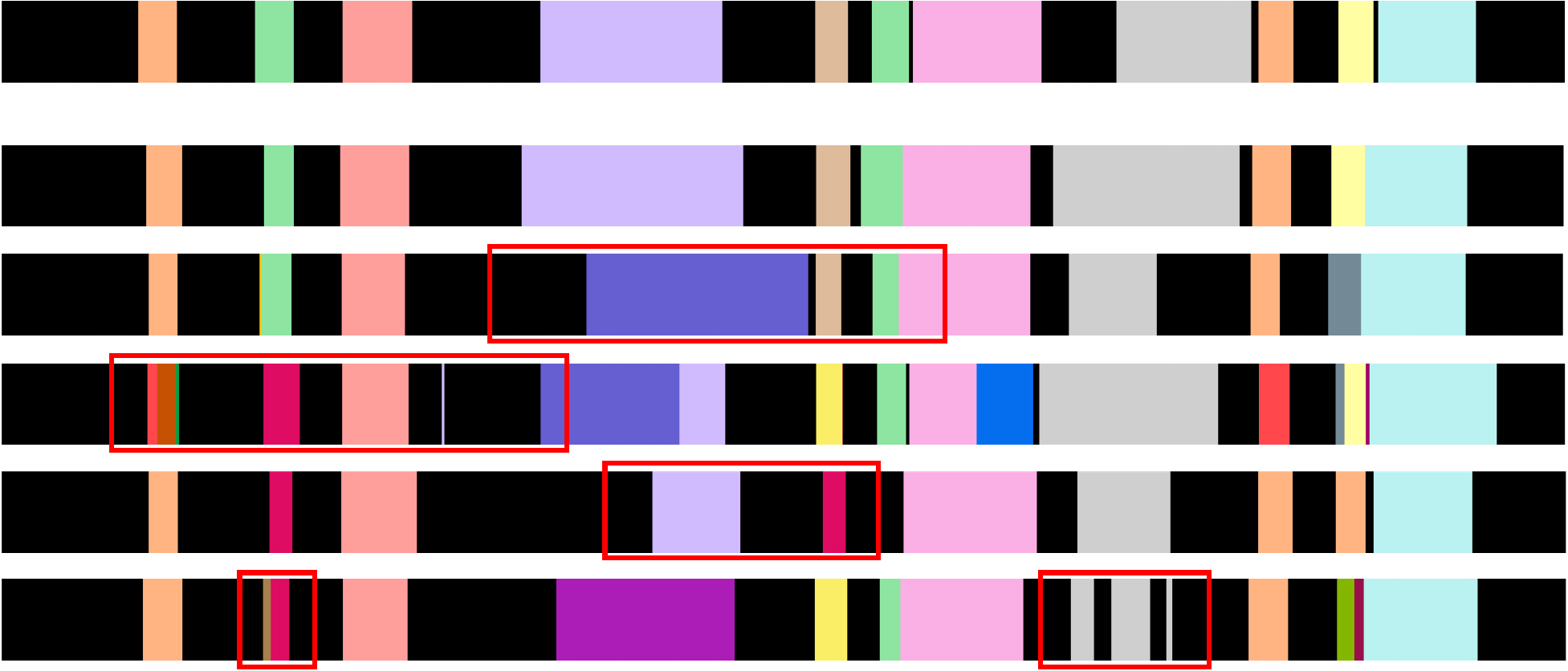}
            \put(-24,38){\footnotesize{Ground Truth}}
            \put(-23,29){\textbf{\footnotesize{\ourMthd~(ours)}}} 
            \put(-18,22){\footnotesize{IDT-GCN}}
            \put(-10.7,15){\footnotesize{BCN}}
            \put(-12.5,8){\footnotesize{ASRF}}
            \put(-12.5,1){\footnotesize{ETSN}}
            \put(35,-5){\footnotesize{(a) MCFS-130}}
        \end{overpic}
    \end{minipage}
    \begin{minipage}[t]{0.49\linewidth}
        \flushright
        \begin{overpic}[width=0.8\textwidth]{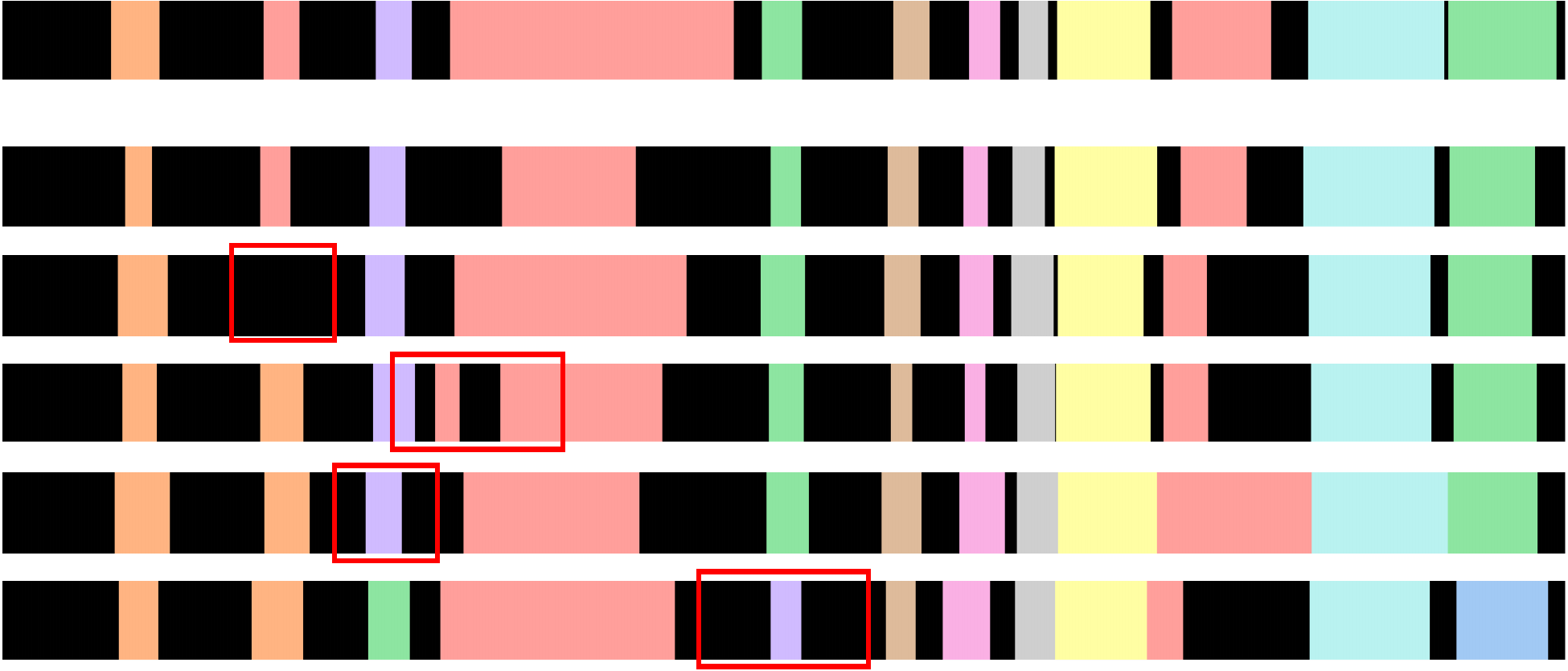}
            \put(-24,38){\footnotesize{Ground Truth}}
            \put(-23,29){\textbf{\footnotesize{\ourMthd~(ours)}}} 
            \put(-18,22){\footnotesize{IDT-GCN}}
            \put(-10.8,15){\footnotesize{BCN}}
            \put(-12.5,8){\footnotesize{ASRF}}
            \put(-12.5,1){\footnotesize{ETSN}}
            \put(35,-5){\footnotesize{(b) MCFS-22}}
        \end{overpic}
    \end{minipage}
    \\
    \vspace{.5cm}
    \\
    \begin{minipage}[t]{0.49\linewidth}
        \flushright
        \begin{overpic}[width=0.8\textwidth]{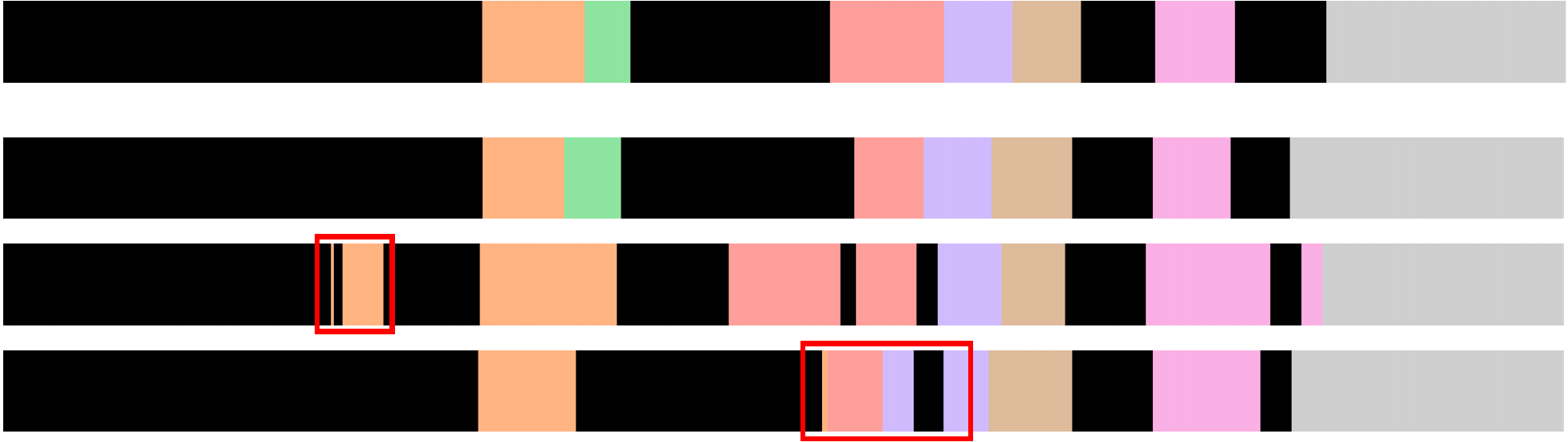}
            \put(-24,24){\footnotesize{Ground Truth}}
            \put(-23,15){\textbf{\footnotesize{\ourMthd~(ours)}}} 
            \put(-17.5,8){\footnotesize{MS-GCN}}
            \put(-17.1,1){\footnotesize{MS-TCN}}
            \put(30,-5){\footnotesize{(c) PKU-MMD (x-sub)}}
        \end{overpic}
    \end{minipage}
    \begin{minipage}[t]{0.49\linewidth}
        \flushright
        \begin{overpic}[width=0.8\textwidth]{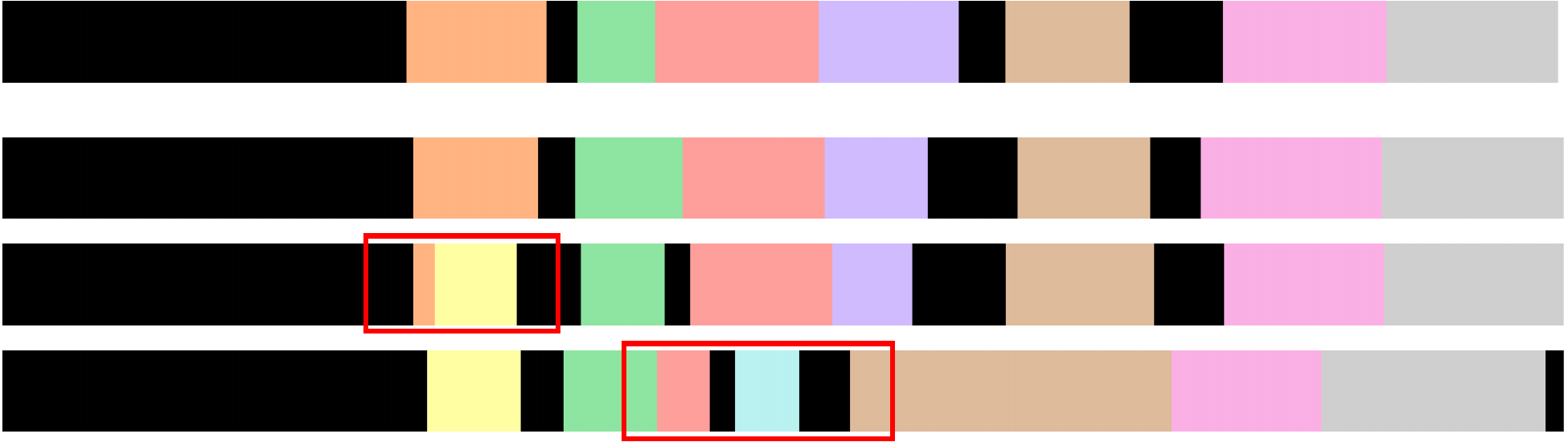}
            \put(-24,24){\footnotesize{Ground Truth}}
            \put(-23,15){\textbf{\footnotesize{\ourMthd~(ours)}}} 
            \put(-17.5,8){\footnotesize{MS-GCN}}
            \put(-17.1,1){\footnotesize{MS-TCN}}
            \put(30,-5){\footnotesize{(d) PKU-MMD (x-view)}}
        \end{overpic}
    \end{minipage}
   \vspace{.5cm}
  \caption{The qualitative results of action segmentation on MCFS-130, MCFS-22, and PKU-MMD datasets.
  The segmentation visualization is in the temporal order of actions (from left to right).
  The different colors represent different action categories.
  The \textcolor{red}{red} box indicates that the models wrongly detect some actions or produce many over-segmentation errors.
  }
  \label{Visualization-MCFS}
\end{figure*}

%% file: figs/Groups-sub-features.tex
\begin{figure}[!t]
  \flushright
    \begin{overpic}[width=0.45\textwidth]{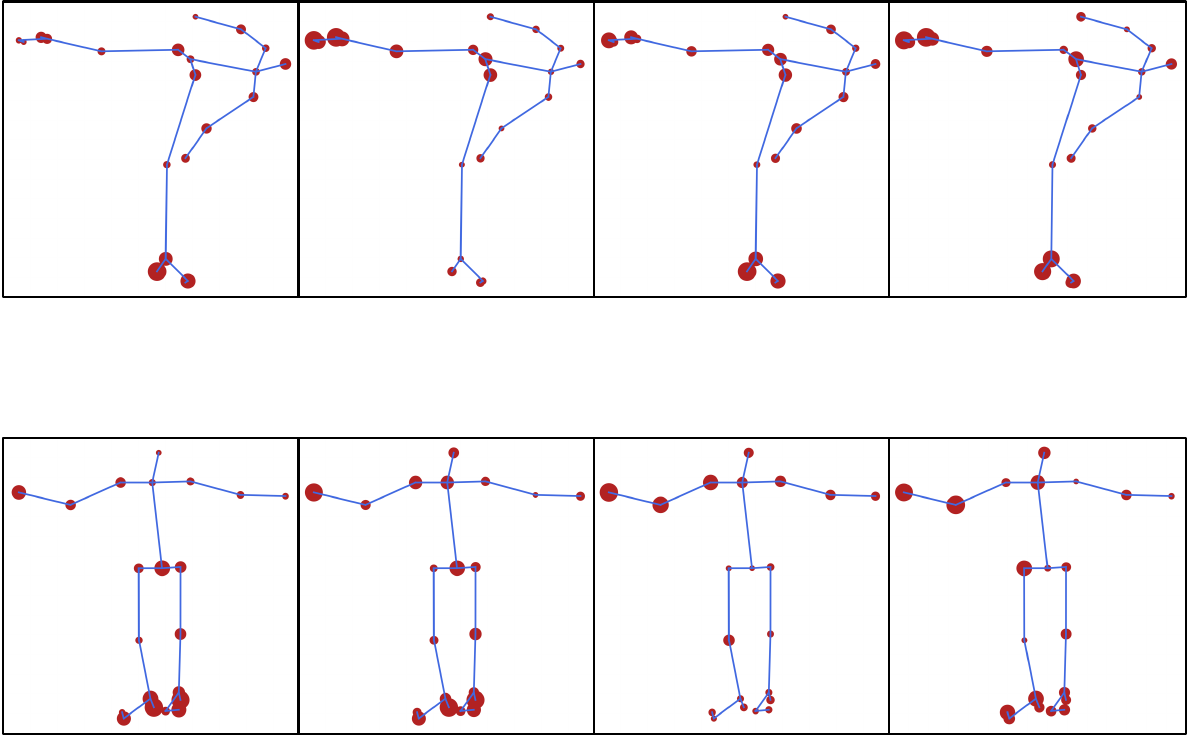}
    \put(6,33){\footnotesize {Group[0]}}
    \put(30,33){\footnotesize {Group[1]}}
    \put(55,33){\footnotesize {Group[3]}}
    \put(80,33){\footnotesize {Group[4]}}
    \put(-7,47){(a)}
    % % 
    \put(6,-3){\footnotesize {Group[0]}}
    \put(30,-3){\footnotesize {Group[1]}}
    \put(55,-3){\footnotesize {Group[3]}}
    \put(80,-3){\footnotesize {Group[4]}}
    \put(-7,10){(b)}
    \end{overpic}
   \vspace{.5cm}
  \caption{The activation values~(proportional to the size of the red dot) 
  of different groups of spatial sub-features.
  The larger the red dot is, 
  the corresponding joint 
  play a more important role in spatio-temporal modeling. 
  (a) and (b) show the sub-features in two different actions, \ie Toeloop\_Toeloop and StepSequence, respectively.
  }
  \label{Visualization-Groups-sub-features}
\end{figure}
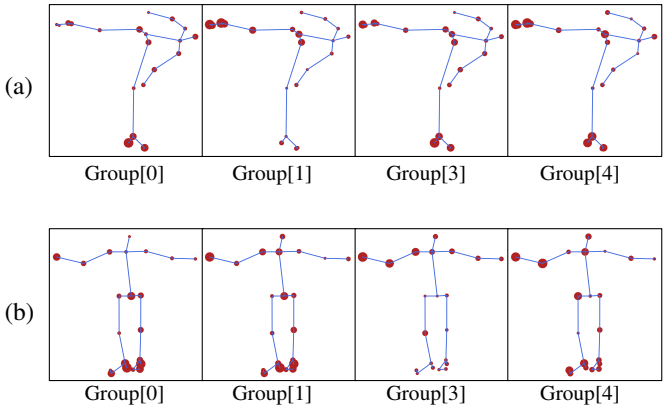

%% file: tabs/X-SSM-STI.tex
\begin{table}[!t]
  \centering
  \setlength{\abovecaptionskip}{0pt}
  \caption{Comparison of various spatio-temporal interaction approaches 
  on the MCFS-130 dataset (split \#1).
  SSF~(\ding{52}) denotes that spatial sub-features are employed. 
  The summation and cross-attention are different interaction manners in 
  the proposed decoupled spatio-temporal interaction~(STI). 
  % and STI means using th different interactive operations. 
  }
  \setlength{\tabcolsep}{1.2mm}
  \subfloat[Adopting TCN in joint-decoupled temporal modeling.]{
    \begin{tabular}{cccccccc}
      \toprule
      SSF &STI  & F1$@10$~$\uparrow$ & F1$@25$~$\uparrow$ & F1$@50$~$\uparrow$ &Edit~$\uparrow$ &Acc~$\uparrow$\\
      \midrule
      % \multicolumn{2}{l}{Baseline} & 71.2& 67.5 & 59.1& 70.6& 68.8\\ \midrule
      \ding{55} &summation& 69.3 & 65.9 & 56.8 & 68.9 & 68.6 \\
      \ding{52} &summation & 72.5 & 69.0 & 60.4 & 71.5 & 69.0 \\
      \ding{55} &cross-attention & 73.2 & 69.0 & 59.5 & 71.8 & 69.5 \\
      \ding{52} &cross-attention & \textbf{74.3} & \textbf{71.5} & \textbf{61.9} & \textbf{73.7} & \textbf{69.9} \\
    \toprule
    \end{tabular}
    }
    \\
    \subfloat[Adopting transformer in joint-decoupled temporal modeling.]{
  \begin{tabular}{cccccccc}
    \toprule
    SSF &STI& F1$@10$~$\uparrow$ & F1$@25$~$\uparrow$ & F1$@50$~$\uparrow$ &Edit~$\uparrow$ &Acc~$\uparrow$\\
      \midrule
 % \multicolumn{2}{l}{Baseline} & 72.5 & 69.4 & 59.7 & 73.0 & 69.4 \\ \midrule
  \ding{55} &summation & 72.6 & 69.3 & 60.2 & 73.3 & 69.9 \\
  \ding{52} &summation & 73.1 & 69.8 & 60.8 & 72.1 & 69.2 \\
  \ding{55} &cross-attention & 74.2 & 70.1 & 61.7 & 74.3 & 69.9 \\
  \ding{52} &cross-attention & \textbf{76.1} & \textbf{71.8} & \textbf{62.1} &\textbf{76.3} & \textbf{70.6} \\
  \toprule
    \end{tabular}
    }
\label{SSM-STI}
\end{table}

%% file: figs/13-Visualization-Compared-Tsne.tex
\begin{figure}[!t]
  \begin{minipage}[t]{0.98\linewidth}
    \flushright
    \begin{overpic}[width=0.82\textwidth]{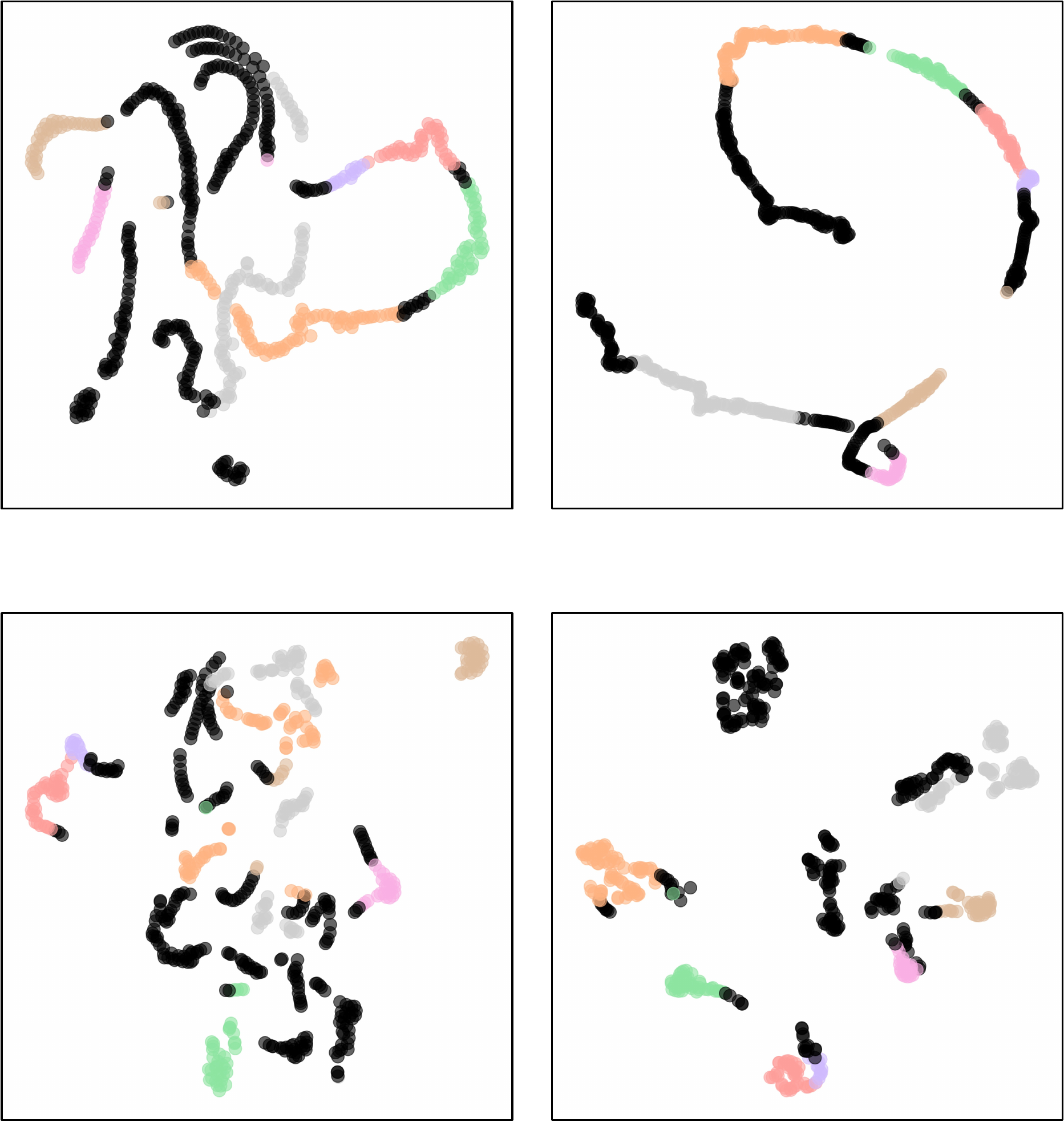}
        \put(-17,73){\footnotesize{MS-GCN}}
        \put(-21.5,20){\footnotesize{\ourMthd~(Ours)}}
        \put(16,50){\small{layer $=5$}}
        \put(61,50){\small{layer $=10$}}
        \put(16,-5){\small{layer $=5$}}
        \put(61,-5){\small{layer $=10$}}
        \put(35,-11){\small{(a) t-SNE}}
    \end{overpic}
\end{minipage}
\\
\vspace{1.0cm}
\\
   \begin{minipage}[t]{0.98\linewidth}
    \flushright
    \begin{overpic}[width=0.82\textwidth]{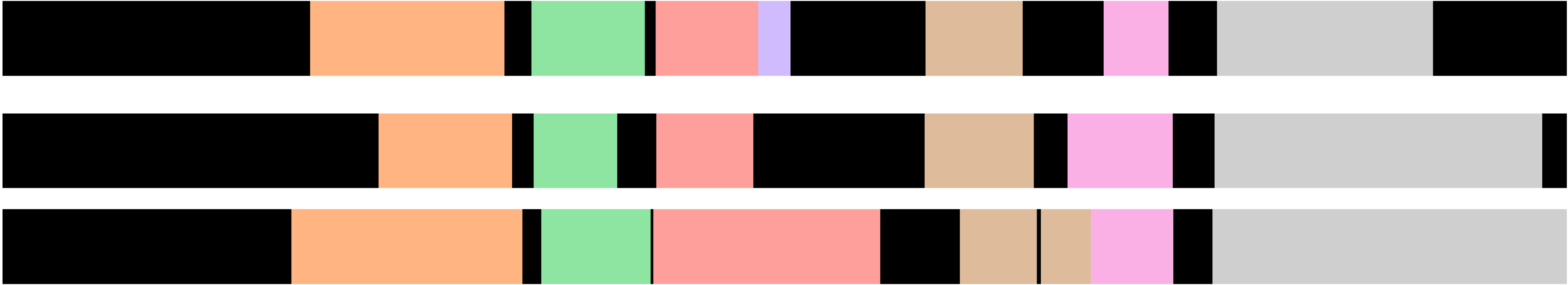}
        \put(-23,14){\footnotesize{Ground Truth}}
        \put(-22,6.8){\footnotesize{\ourMthd~(Ours)}}
        \put(-17,0){\footnotesize{MS-GCN}}
        \put(35,-7){\small{(b) Predictions}}
    \end{overpic}
\end{minipage}   
   \vspace{.5cm}
  \caption{The (a) t-SNE visualization and (b) corresponding predictions 
  from different layers of MS-GCN\cite{filtjens2022skeleton} and 
  the proposed \ourMthd.
  % 
  % (a) The feature embedding of the different layers in a 2D space using t-SNE \cite{van2008visualizing}.
  % 
  In (a), 
  the features output by the temporal modeling layer are visualized and 
  different colors indicate the different categories in the PKU-MMD (x-sub).
  % 
  % The dots represent the last feature embedding of the encoder that matches the ground truth annotations.
  % 
  % $l$ represents the layer of the network.
  % 
  In (b), the three rows show the ground-truth labels, the predictions from \ourMthd, and the predictions from MS-GCN.
  }
  \label{Visualization-Compared-Tsne}
\end{figure}

%% file: figs/8-Statistic-Joint-Speed.tex
\begin{figure}[!t]
  \centering
   \begin{overpic}[width=0.45\textwidth]{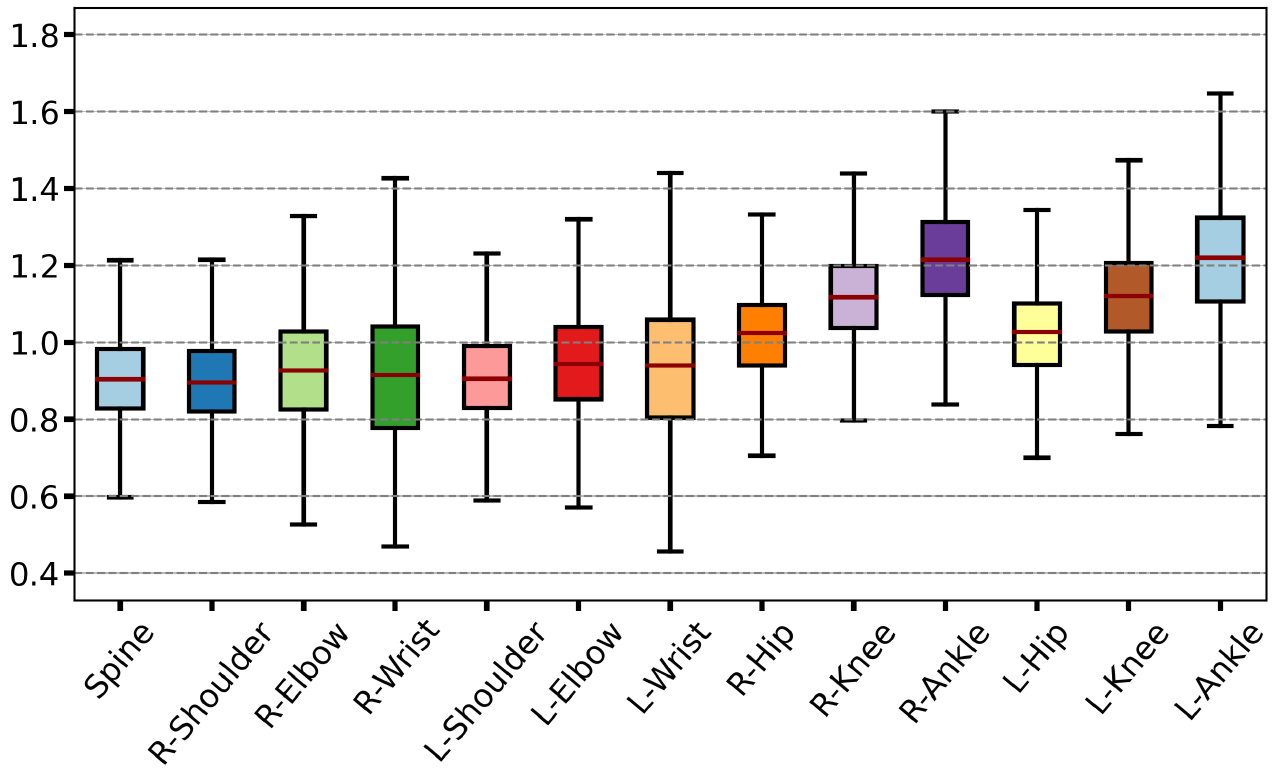} 
    \put(40,-2){\footnotesize{Selected Joints}}
    \put(-3,25){\rotatebox{90} {\footnotesize{Values of Speed}}}
   \end{overpic}
    \vspace{.2cm}
  \caption{Motion speeds of different joints. 
  We count the speeds of joint by all samples in the MCFS-130 dataset.
  Different boxes represent the speed range of different joints: 
  1) the red line inside each box denotes the median value; 
  2) boxes boundaries means the first quartile and the third quartile of the data, respectively; 
  3) upper and lower whiskers indicate the max value and min value, respectively.
  }
  \label{Statistic-Joint-Speed}
\end{figure}

%% file: tabs/VII-JTM-JSTM.tex
\begin{table}[!t]
  \centering
  \setlength{\abovecaptionskip}{0pt}
  \caption{Comparison of joint-decoupled and joint-shared weights.
    JWTM defined by \secref{sec:exp-JTM} means the module that uses shared weights to model all joints.}
  \setlength{\tabcolsep}{1.0mm}
  \subfloat[Comparison on the MCFS-22 dataset~(split \#1).]{
        \begin{tabular}{lccccccc}
            \toprule
            \multicolumn{2}{c}{Model}& F1$@10$~$\uparrow$ & F1$@25$~$\uparrow$ & F1$@50$~$\uparrow$ & Edit~$\uparrow$& Acc~$\uparrow$ \\
            \midrule
            \multirow{3}{*}{\makecell[c]{TCN}} 
            & 2D-Conv & 85.3 & 81.6 & 70.8 & 80.9 & 78.1  \\
            & JWTM & 85.7 & 82.0 & 71.1 & 81.6 & 77.4  \\
            & JTM & \textbf{87.2} &  \textbf{83.6} &  \textbf{72.8} &  \textbf{82.6} &  \textbf{78.1}  \\
            \midrule
            \multirow{2}{*}{\makecell[c]{Transformer}} 
            & JWTM & 80.6 & 75.9 & 65.6 & 78.7 & 75.3   \\ 
            & JTM &  \textbf{86.7} &  \textbf{83.6} &  \textbf{73.9} &  \textbf{83.7} &  \textbf{79.1}  \\
            \bottomrule
        \end{tabular}
    }
    \\
    \subfloat[Comparison on the MCFS-130 dataset~(split \#1).]{
        \begin{tabular}{lccccccc}
            \toprule
            \multicolumn{2}{c}{Model}& F1$@10$~$\uparrow$ & F1$@25$~$\uparrow$ & F1$@50$~$\uparrow$ & Edit~$\uparrow$& Acc~$\uparrow$ \\
            \midrule
            \multirow{3}{*}{\makecell[c]{TCN}} 
            & 2D-Conv & 68.8 & 65.3 & 56.1 & 67.0 & 67.0 \\
            & JWTM & 69.4 & 65.8 & 57.1 & 68.6 & 67.9 \\
            & JTM &  \textbf{71.2} &  \textbf{67.5} &  \textbf{59.1} &  \textbf{70.6} &  \textbf{68.8} \\
            \midrule
            \multirow{2}{*}{\makecell[c]{Transformer}} 
            & JWTM & 71.4 & 66.5 & 58.0 & 71.6 & 68.1\\
            & JTM &  \textbf{72.5} &  \textbf{69.4} &  \textbf{59.7} &  \textbf{73.0} &  \textbf{69.4} \\
            \bottomrule
        \end{tabular}
    }
\label{JTM-JSTM}
\end{table}

%% file: figs/9-Visualization-JTM-JSTM-mean.tex
\begin{figure}[!t]
  \centering
    \begin{overpic}[width=0.48\textwidth]{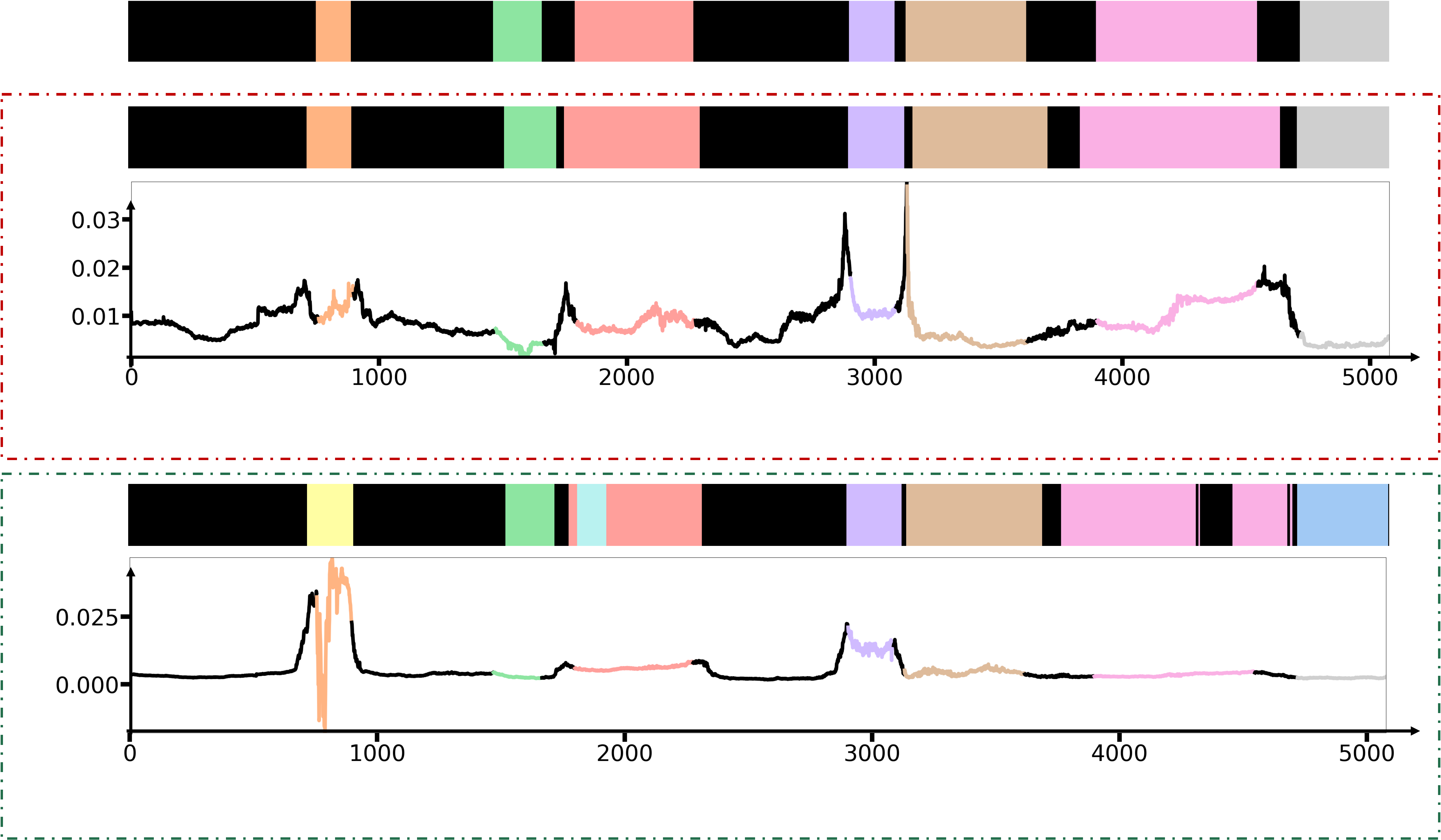}
    % JTM
    \put(2,55){\small {GT}}
    \put(0.5,47){\footnotesize {Pred.}}
    \put(0.5,33){\rotatebox{90} {\footnotesize {Features}}}
    \put(97,29){{\scriptsize {$T$}}}
    \put(18,28.1){\small{(a) Joint-decoupled Temporal Modeling (ours)}}
    % JSTM
    \put(0.5,21){\footnotesize {Pred.}}
    \put(0.5,7){\rotatebox{90} {\footnotesize {Features}}}
    \put(97,3){{\scriptsize {$T$}}}
    \put(20,1.1){\small {(b) Joint-shared Weight Temporal Modeling.}}
    \end{overpic}
   % \vspace{.5cm}
  \caption{The visualization of predictions and temporal features output by JTM and JWTM. 
  We take a sequence from the MCFS-22 dataset as an example.
  The dimension of the temporal feature is $C^t \times V$, and the horizontal axis represents the temporal order of actions (from left to right).
  About visualization, 
  we average the temporal features along channels, 
  and different colors represent different categories of action.
  }
  \label{Visualization-JTM-JSTM-mean}
\end{figure}
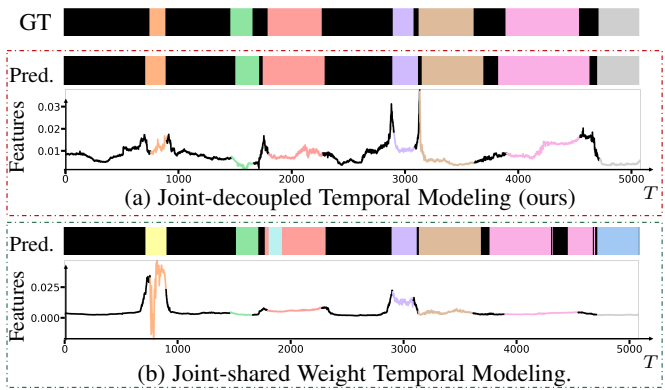

%% file: tabs/IX-SSM-Transformation_functions.tex
\begin{table}[t]
  \centering
  \setlength{\abovecaptionskip}{0pt}
  \caption{The effect of different feature-transform functions on the MCFS-22 dataset (split \#1).
  }
\setlength{\tabcolsep}{1.5mm}{
  \begin{tabular}{lccccccc}
  \toprule 
 &$\mathcal {R}$  & F1$@10$~$\uparrow$ & F1$@25$~$\uparrow$ & F1$@50$~$\uparrow$ & Edit~$\uparrow$ & Acc~$\uparrow$ \\
      \midrule
      %   \multicolumn{1}{l}{JSTM} \\
      % &Conv & 80.6 & 75.9 & 65.6 & 78.7 & 75.3   \\ 
      % \hline
      % JTM\\
      &Avg pooling& 86.4    & 82.7    & 73.2    & 82.8 & 78.6   \\
      &Max pooling& 84.6    & 81.1    & 71.0    & 80.9 & 78.3   \\
      &Convoluation& \textbf{86.7}& \textbf{83.6}& \textbf{73.9}& \textbf{83.7} & \textbf{79.1}   \\
  \bottomrule 
\end{tabular}}
\label{transformation_functions}
\end{table}

%% file: tabs/XII-JTM-layers.tex
\begin{table}[!t]
  \centering
  \setlength{\abovecaptionskip}{0pt}
  \caption{The effect of the number of the JTM layers ($L_y$) for temporal modeling on MCFS-130 dataset (split \#1).
  }
\setlength{\tabcolsep}{2.5mm}{
  \begin{tabular}{cccccc}
  \toprule 
   $L_y$ & F1$@10$~$\uparrow$ & F1$@25$~$\uparrow$ & F1$@50$~$\uparrow$ & Edit~$\uparrow$ & Acc~$\uparrow$ \\
  \midrule
  8& 72.0 & 68.7 & 58.5 & 74.0 & 69.7 \\ 
  9& 74.1 & 70.9 & 61.6 & 73.5 & 70.3 \\
  10&\textbf{76.3} & \textbf{73.1} & \textbf{63.4} & \textbf{76.8} & \textbf{70.5} \\
  11& 75.2 & 71.3 & 61.2 & 75.5 & 69.5 \\
  12& 74.8 & 72.0 & 60.8 & 75.3 & 69.6 \\
  \bottomrule 
\end{tabular}}
\label{JTM-layers}
\end{table}

%% file: tabs/XIII-STI-layers.tex
\begin{table}[!t]
  \centering
  \setlength{\tabcolsep}{2.5mm}
  \setlength{\abovecaptionskip}{0pt}
  \caption{The effect of the number of the interaction~(DSTI) layers ($L_c$) 
  on MCFS-130 dataset (split \#1).}
  \begin{tabular}{ccccccc}
  \bottomrule
    $L_c$ & F1$@10$~$\uparrow$ & F1$@25$~$\uparrow$ & F1$@50$~$\uparrow$ & Edit~$\uparrow$ & Acc~$\uparrow$ \\
  \midrule
    1 & 72.6 & 69.2 & 60.7 & 73.1 & 69.0 \\
    2 & 73.3 & 70.0 & 60.9 & 73.1 & 69.9 \\
    3 & 74.2 & 70.6 & 61.5 & 73.3 & 70.4 \\
    4 & 74.3 & 70.1 & 62.0 & 74.4 & 69.5 \\
    5 & 74.8 & 71.0 & 61.1 & 74.4 & 70.0 \\
    6 & 74.8 & 71.0 & 62.7 & 74.3 & 70.4 \\
    7 & 75.1 & 72.0 & 62.1 & 75.0 & \textbf{70.7} \\
    8 & 75.0 & 71.7 & 62.8 & 75.0 & \textbf{70.7} \\
    9 & \textbf{76.3} & \textbf{73.1} & \textbf{63.4} & \textbf{76.8} & 70.5\\
  \toprule 
  \end{tabular}
\label{STI-layers}
\end{table}

%% file: figs/14-STI-CrossAttention-Map.tex
\begin{figure}[t]
  \centering
    \begin{overpic}[width=0.45\textwidth]{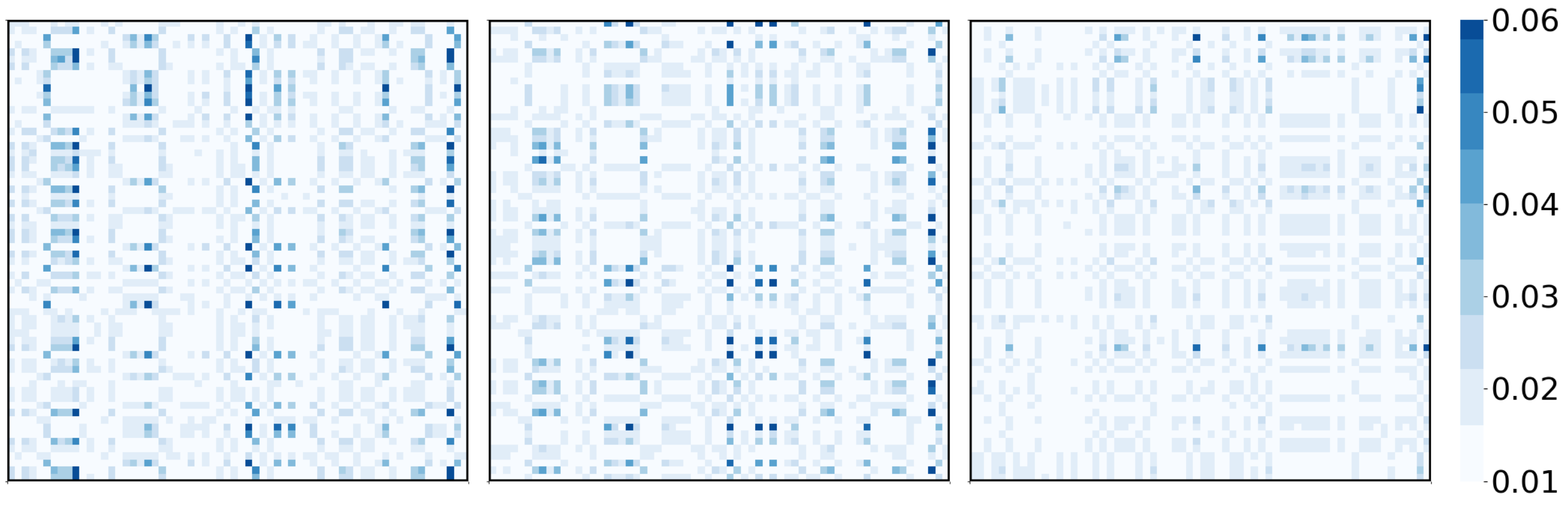}
        \put(9.5,-2){\footnotesize layer $=1$}
        \put(39.5,-2){\footnotesize layer $=5$}
        \put(69.5,-2){\footnotesize layer $=9$}
    \end{overpic}
   % \vspace{.2cm}
  \caption{The visualization of cross-attention weights of different interaction~(DSTI) layers.
  }
  \label{STI-CrossAttention-Map}
\end{figure}

%% file: J-STFormer.bbl
\begin{thebibliography}{10}
\providecommand{\url}[1]{#1}
\csname url@samestyle\endcsname
\providecommand{\newblock}{\relax}
\providecommand{\bibinfo}[2]{#2}
\providecommand{\BIBentrySTDinterwordspacing}{\spaceskip=0pt\relax}
\providecommand{\BIBentryALTinterwordstretchfactor}{4}
\providecommand{\BIBentryALTinterwordspacing}{\spaceskip=\fontdimen2\font plus
\BIBentryALTinterwordstretchfactor\fontdimen3\font minus
  \fontdimen4\font\relax}
\providecommand{\BIBforeignlanguage}[2]{{%
\expandafter\ifx\csname l@#1\endcsname\relax
\typeout{** WARNING: IEEEtran.bst: No hyphenation pattern has been}%
\typeout{** loaded for the language `#1'. Using the pattern for}%
\typeout{** the default language instead.}%
\else
\language=\csname l@#1\endcsname
\fi
#2}}
\providecommand{\BIBdecl}{\relax}
\BIBdecl

\bibitem{luo2019capturing}
X.~Luo, H.~Li, X.~Yang, Y.~Yu, and D.~Cao, ``Capturing and understanding
  workers’ activities in far-field surveillance videos with deep action
  recognition and bayesian nonparametric learning,'' \emph{Comput-aided Civ.
  Inf.}, vol.~34, no.~4, pp. 333--351, 2019.

\bibitem{son2019detection}
H.~Son, H.~Choi, H.~Seong, and C.~Kim, ``Detection of construction workers
  under varying poses and changing background in image sequences via very deep
  residual networks,'' \emph{Automat. Constr.}, vol.~99, pp. 27--38, 2019.

\bibitem{filtjens2020data}
B.~Filtjens, A.~Nieuwboer, N.~D’cruz, J.~Spildooren, P.~Slaets, and
  B.~Vanrumste, ``A data-driven approach for detecting gait events during
  turning in people with parkinson's disease and freezing of gait,'' \emph{Gait
  \& Posture}, vol.~80, pp. 130--136, 2020.

\bibitem{kidzinski2019automatic}
{\L}.~Kidzi{\'n}ski, S.~Delp, and M.~Schwartz, ``Automatic real-time gait event
  detection in children using deep neural networks,'' \emph{PLoS ONE}, vol.~14,
  no.~1, p. e0211466, 2019.

\bibitem{kenney2009interactive}
J.~Kenney, T.~Buckley, and O.~Brock, ``Interactive segmentation for
  manipulation in unstructured environments,'' in \emph{ICRA}.\hskip 1em plus
  0.5em minus 0.4em\relax IEEE, 2009, pp. 1377--1382.

\bibitem{siam2019video}
M.~Siam, C.~Jiang, S.~Lu, L.~Petrich, M.~Gamal, M.~Elhoseiny, and M.~Jagersand,
  ``Video object segmentation using teacher-student adaptation in a human robot
  interaction (hri) setting,'' in \emph{ICRA}.\hskip 1em plus 0.5em minus
  0.4em\relax IEEE, 2019, pp. 50--56.

\bibitem{sudha2017approaches}
M.~Sudha, K.~Sriraghav, S.~G. Jacob, S.~Manisha \emph{et~al.}, ``Approaches and
  applications of virtual reality and gesture recognition: A review,''
  \emph{IJCAI}, vol.~8, no.~4, pp. 1--18, 2017.

\bibitem{biswas2014kinect}
A.~Biswas, S.~Dutta, N.~Dey, and A.~T. Azar, ``A kinect-less augmented reality
  approach to real-time tag-less virtual trial room simulation,''
  \emph{IJSSMET}, vol.~5, no.~4, pp. 13--28, 2014.

\bibitem{liu2017enhanced}
M.~Liu, H.~Liu, and C.~Chen, ``Enhanced skeleton visualization for view
  invariant human action recognition,'' \emph{Pattern Recognit.}, vol.~68, pp.
  346--362, 2017.

\bibitem{song2022constructing}
Y.-F. Song, Z.~Zhang, C.~Shan, and L.~Wang, ``Constructing stronger and faster
  baselines for skeleton-based action recognition,'' \emph{IEEE Trans. Pattern
  Anal. Mach. Intell.}, vol.~45, no.~2, pp. 1474--1488, 2022.

\bibitem{cheng2021extremely}
K.~Cheng, Y.~Zhang, X.~He, J.~Cheng, and H.~Lu, ``Extremely lightweight
  skeleton-based action recognition with shiftgcn++,'' \emph{IEEE Trans. Image
  Process.}, vol.~30, pp. 7333--7348, 2021.

\bibitem{carreira2017quo}
J.~Carreira and A.~Zisserman, ``Quo vadis, action recognition? a new model and
  the kinetics dataset,'' in \emph{IEEE Conf. Comput. Vis. Pattern Recog.},
  2017, pp. 6299--6308.

\bibitem{simonyan2014two}
K.~Simonyan and A.~Zisserman, ``Two-stream convolutional networks for action
  recognition in videos,'' \emph{Adv. Neural Inform. Process. Syst.}, vol.~27,
  2014.

\bibitem{wang2018appearance}
L.~Wang, W.~Li, W.~Li, and L.~Van~Gool, ``Appearance-and-relation networks for
  video classification,'' in \emph{IEEE Conf. Comput. Vis. Pattern Recog.},
  2018, pp. 1430--1439.

\bibitem{wang2015action}
L.~Wang, Y.~Qiao, and X.~Tang, ``Action recognition with trajectory-pooled
  deep-convolutional descriptors,'' in \emph{IEEE Conf. Comput. Vis. Pattern
  Recog.}, 2015, pp. 4305--4314.

\bibitem{wang2016temporal}
L.~Wang, Y.~Xiong, Z.~Wang, Y.~Qiao, D.~Lin, X.~Tang, and L.~Van~Gool,
  ``Temporal segment networks: Towards good practices for deep action
  recognition,'' in \emph{Eur. Conf. Comput. Vis.}\hskip 1em plus 0.5em minus
  0.4em\relax Springer, 2016, pp. 20--36.

\bibitem{liu2022video}
Z.~Liu, J.~Ning, Y.~Cao, Y.~Wei, Z.~Zhang, S.~Lin, and H.~Hu, ``Video swin
  transformer,'' in \emph{IEEE Conf. Comput. Vis. Pattern Recog.}, 2022, pp.
  3202--3211.

\bibitem{arnab2021vivit}
A.~Arnab, M.~Dehghani, G.~Heigold, C.~Sun, M.~Lu{\v{c}}i{\'c}, and C.~Schmid,
  ``Vivit: A video vision transformer,'' in \emph{Int. Conf. Comput. Vis.},
  2021, pp. 6836--6846.

\bibitem{Wang_2023_CVPR}
L.~Wang, B.~Huang, Z.~Zhao, Z.~Tong, Y.~He, Y.~Wang, Y.~Wang, and Y.~Qiao,
  ``Videomae v2: Scaling video masked autoencoders with dual masking,'' in
  \emph{IEEE Conf. Comput. Vis. Pattern Recog.}, June 2023, pp.
  14\,549--14\,560.

\bibitem{filtjens2022skeleton}
B.~Filtjens, B.~Vanrumste, and P.~Slaets, ``Skeleton-based action segmentation
  with multi-stage spatial-temporal graph convolutional neural networks,''
  \emph{IEEE Trans. Emerg. Top. Com.}, 2022.

\bibitem{xu2023efficient}
L.~Xu, Q.~Wang, X.~Lin, and L.~Yuan, ``An efficient framework for few-shot
  skeleton-based temporal action segmentation,'' \emph{Comput. Vis. Image
  Und.}, vol. 232, p. 103707, 2023.

\bibitem{liu2022spatial}
K.~Liu, Y.~Li, Y.~Xu, S.~Liu, and S.~Liu, ``Spatial focus attention for
  fine-grained skeleton-based action tasks,'' \emph{IEEE Sign. Process.
  Letters}, vol.~29, pp. 1883--1887, 2022.

\bibitem{li2023involving}
Y.-H. Li, K.-Y. Liu, S.-L. Liu, L.~Feng, and H.~Qiao, ``Involving distinguished
  temporal graph convolutional networks for skeleton-based temporal action
  segmentation,'' \emph{IEEE Trans. Circuit Syst. Video Technol.}, 2023.

\bibitem{fagcn2021}
D.~Bo, X.~Wang, C.~Shi, and H.~Shen, ``Beyond low-frequency information in
  graph convolutional networks,'' in \emph{AAAI}.\hskip 1em plus 0.5em minus
  0.4em\relax {AAAI} Press, 2021.

\bibitem{pang2023skeleton}
C.~Pang, X.~Lu, and L.~Lyu, ``Skeleton-based action recognition through
  contrasting two-stream spatial-temporal networks,'' \emph{IEEE Trans.
  Multimedia}, 2023.

\bibitem{liu2021temporal}
S.~Liu, A.~Zhang, Y.~Li, J.~Zhou, L.~Xu, Z.~Dong, and R.~Zhang, ``Temporal
  segmentation of fine-grained semantic action: A motion-centered figure
  skating dataset,'' in \emph{AAAI}, vol.~35, no.~3, 2021, pp. 2163--2171.

\bibitem{liu2017pku}
C.~Liu, Y.~Hu, Y.~Li, S.~Song, and J.~Liu, ``Pku-mmd: A large scale benchmark
  for skeleton-based human action understanding,'' in \emph{ACM VASCCW}, 2017,
  pp. 1--8.

\bibitem{niemann2020lara}
F.~Niemann, C.~Reining, F.~Moya~Rueda, N.~R. Nair, J.~A. Steffens, G.~A. Fink,
  and M.~Ten~Hompel, ``Lara: Creating a dataset for human activity recognition
  in logistics using semantic attributes,'' \emph{Sensors}, vol.~20, no.~15, p.
  4083, 2020.

\bibitem{Carreira_2017_CVPR}
J.~Carreira and A.~Zisserman, ``Quo vadis, action recognition? a new model and
  the kinetics dataset,'' in \emph{IEEE Conf. Comput. Vis. Pattern Recog.},
  July 2017.

\bibitem{ding2017tricornet}
L.~Ding and C.~Xu, ``Tricornet: A hybrid temporal convolutional and recurrent
  network for video action segmentation,'' \emph{arXiv preprint
  arXiv:1705.07818}, 2017.

\bibitem{Singh_2016_CVPR}
B.~Singh, T.~K. Marks, M.~Jones, O.~Tuzel, and M.~Shao, ``A multi-stream
  bi-directional recurrent neural network for fine-grained action detection,''
  in \emph{IEEE Conf. Comput. Vis. Pattern Recog.}, June 2016.

\bibitem{lea2017temporal}
C.~Lea, M.~D. Flynn, R.~Vidal, A.~Reiter, and G.~D. Hager, ``Temporal
  convolutional networks for action segmentation and detection,'' in \emph{IEEE
  Conf. Comput. Vis. Pattern Recog.}, 2017, pp. 156--165.

\bibitem{li2021efficient}
Y.~Li, Z.~Dong, K.~Liu, L.~Feng, L.~Hu, J.~Zhu, L.~Xu, S.~Liu \emph{et~al.},
  ``Efficient two-step networks for temporal action segmentation,''
  \emph{Neurocomputing}, vol. 454, pp. 373--381, 2021.

\bibitem{farha2019ms}
Y.~A. Farha and J.~Gall, ``Ms-tcn: Multi-stage temporal convolutional network
  for action segmentation,'' in \emph{IEEE Conf. Comput. Vis. Pattern Recog.},
  2019, pp. 3575--3584.

\bibitem{li2020ms}
S.~Li, Y.~A. Farha, Y.~Liu, M.-M. Cheng, and J.~Gall, ``Ms-tcn++: Multi-stage
  temporal convolutional network for action segmentation,'' \emph{IEEE Trans.
  Pattern Anal. Mach. Intell.}, vol.~45, no.~6, pp. 6647--6658, 2023.

\bibitem{Gao_2021_CVPR}
S.-H. Gao, Q.~Han, Z.-Y. Li, P.~Peng, L.~Wang, and M.-M. Cheng, ``Global2local:
  Efficient structure search for video action segmentation,'' in \emph{IEEE
  Conf. Comput. Vis. Pattern Recog.}, June 2021, pp. 16\,805--16\,814.

\bibitem{gao2022rf}
S.~Gao, Z.-Y. Li, Q.~Han, M.-M. Cheng, and L.~Wang, ``Rf-next: Efficient
  receptive field search for convolutional neural networks,'' \emph{IEEE Trans.
  Pattern Anal. Mach. Intell.}, vol.~45, no.~3, pp. 2984--3002, 2022.

\bibitem{chinayi_ASformer}
F.~Yi, H.~Wen, and T.~Jiang, ``Asformer: Transformer for action segmentation,''
  in \emph{Brit. Mach. Vis. Conf.}, 2021.

\bibitem{behrmann2022unified}
N.~Behrmann, S.~A. Golestaneh, Z.~Kolter, J.~Gall, and M.~Noroozi, ``Unified
  fully and timestamp supervised temporal action segmentation via sequence to
  sequence translation,'' in \emph{Eur. Conf. Comput. Vis.}\hskip 1em plus
  0.5em minus 0.4em\relax Springer, 2022, pp. 52--68.

\bibitem{ltc2023bahrami}
J.~G. Emad~Bahrami, Gianpiero~Francesca, ``How much temporal long-term context
  is needed for action segmentation?'' in \emph{Int. Conf. Comput. Vis.}, 2023.

\bibitem{wang2020boundary}
Z.-Z. Wang, Z.-T. Gao, L.-M. Wang, Z.-F. Li, and G.-S. Wu, ``Boundary-aware
  cascade networks for temporal action segmentation,'' in \emph{Eur. Conf.
  Comput. Vis.}\hskip 1em plus 0.5em minus 0.4em\relax Springer, 2020.

\bibitem{ishikawa2021alleviating}
Y.~Ishikawa, S.~Kasai, Y.~Aoki, and H.~Kataoka, ``Alleviating over-segmentation
  errors by detecting action boundaries,'' in \emph{WACV}, 2021, pp.
  2322--2331.

\bibitem{ahn2021refining}
H.~Ahn and D.~Lee, ``Refining action segmentation with hierarchical video
  representations,'' in \emph{Int. Conf. Comput. Vis.}, 2021, pp.
  16\,302--16\,310.

\bibitem{li2022bridge}
M.~Li, L.~Chen, Y.~Duan, Z.~Hu, J.~Feng, J.~Zhou, and J.~Lu, ``Bridge-prompt:
  Towards ordinal action understanding in instructional videos,'' in \emph{IEEE
  Conf. Comput. Vis. Pattern Recog.}, June 2022.

\bibitem{liu2023diffusion}
D.~Liu, Q.~Li, A.-D. Dinh, T.~Jiang, M.~Shah, and C.~Xu, ``Diffusion action
  segmentation,'' in \emph{Int. Conf. Comput. Vis.}, 2023.

\bibitem{chen2021multi}
Z.~Chen, S.~Li, B.~Yang, Q.~Li, and H.~Liu, ``Multi-scale spatial temporal
  graph convolutional network for skeleton-based action recognition,'' in
  \emph{AAAI}, vol.~35, no.~2, 2021, pp. 1113--1122.

\bibitem{wang2020learning}
M.~Wang, B.~Ni, and X.~Yang, ``Learning multi-view interactional skeleton graph
  for action recognition,'' \emph{IEEE Trans. Pattern Anal. Mach. Intell.},
  2020.

\bibitem{tian2023stga}
X.~Tian, Y.~Jin, Z.~Zhang, P.~Liu, and X.~Tang, ``Stga-net: Spatial-temporal
  graph attention network for skeleton-based temporal action segmentation,'' in
  \emph{ICMEW}.\hskip 1em plus 0.5em minus 0.4em\relax IEEE, 2023, pp.
  218--223.

\bibitem{hao2021hypergraph}
X.~Hao, J.~Li, Y.~Guo, T.~Jiang, and M.~Yu, ``Hypergraph neural network for
  skeleton-based action recognition,'' \emph{IEEE Trans. Image Process.},
  vol.~30, pp. 2263--2275, 2021.

\bibitem{li2023graph}
S.~Li, X.~He, W.~Song, A.~Hao, and H.~Qin, ``Graph diffusion convolutional
  network for skeleton based semantic recognition of two-person actions,''
  \emph{IEEE Trans. Pattern Anal. Mach. Intell.}, 2023.

\bibitem{hu2019joint}
G.~Hu, B.~Cui, and S.~Yu, ``Joint learning in the spatio-temporal and frequency
  domains for skeleton-based action recognition,'' \emph{IEEE Trans.
  Multimedia}, vol.~22, no.~9, pp. 2207--2220, 2019.

\bibitem{zhang2019view}
P.~Zhang, C.~Lan, J.~Xing, W.~Zeng, J.~Xue, and N.~Zheng, ``View adaptive
  neural networks for high performance skeleton-based human action
  recognition,'' \emph{IEEE Trans. Pattern Anal. Mach. Intell.}, vol.~41,
  no.~8, pp. 1963--1978, 2019.

\bibitem{nowozin2012action}
S.~Nowozin and J.~Shotton, ``Action points: A representation for low-latency
  online human action recognition,'' \emph{Microsoft Research Cambridge, Tech.
  Rep.}, 2012.

\bibitem{sharaf2015real}
A.~Sharaf, M.~Torki, M.~E. Hussein, and M.~El-Saban, ``Real-time multi-scale
  action detection from 3d skeleton data,'' in \emph{WACV}.\hskip 1em plus
  0.5em minus 0.4em\relax IEEE, 2015, pp. 998--1005.

\bibitem{wang2018beyond}
H.~Wang and L.~Wang, ``Beyond joints: Learning representations from primitive
  geometries for skeleton-based action recognition and detection,'' \emph{IEEE
  Trans. Image Process.}, vol.~27, no.~9, pp. 4382--4394, 2018.

\bibitem{korban2023semantics}
M.~Korban and X.~Li, ``Semantics-enhanced early action detection using dynamic
  dilated convolution,'' \emph{Pattern Recognit.}, vol. 140, p. 109595, 2023.

\bibitem{cui2020learning}
Q.~Cui, H.~Sun, and F.~Yang, ``Learning dynamic relationships for 3d human
  motion prediction,'' in \emph{IEEE Conf. Comput. Vis. Pattern Recog.}, 2020,
  pp. 6519--6527.

\bibitem{Li_2020_CVPR}
M.~Li, S.~Chen, Y.~Zhao, Y.~Zhang, Y.~Wang, and Q.~Tian, ``Dynamic multiscale
  graph neural networks for 3d skeleton based human motion prediction,'' in
  \emph{IEEE Conf. Comput. Vis. Pattern Recog.}, June 2020.

\bibitem{Dang_2021_ICCV}
L.~Dang, Y.~Nie, C.~Long, Q.~Zhang, and G.~Li, ``Msr-gcn: Multi-scale residual
  graph convolution networks for human motion prediction,'' in \emph{Int. Conf.
  Comput. Vis.}, October 2021, pp. 11\,467--11\,476.

\bibitem{li2021symbiotic}
M.~Li, S.~Chen, X.~Chen, Y.~Zhang, Y.~Wang, and Q.~Tian, ``Symbiotic graph
  neural networks for 3d skeleton-based human action recognition and motion
  prediction,'' \emph{IEEE Trans. Pattern Anal. Mach. Intell.}, vol.~44, no.~6,
  pp. 3316--3333, 2021.

\bibitem{yan2018spatial}
S.~Yan, Y.~Xiong, and D.~Lin, ``Spatial temporal graph convolutional networks
  for skeleton-based action recognition,'' in \emph{AAAI}, 2018.

\bibitem{Shi_2019_CVPR}
L.~Shi, Y.~Zhang, J.~Cheng, and H.~Lu, ``Two-stream adaptive graph
  convolutional networks for skeleton-based action recognition,'' in \emph{IEEE
  Conf. Comput. Vis. Pattern Recog.}, June 2019.

\bibitem{cheng2020decoupling}
K.~Cheng, Y.~Zhang, C.~Cao, L.~Shi, J.~Cheng, and H.~Lu, ``Decoupling gcn with
  dropgraph module for skeleton-based action recognition,'' in \emph{Eur. Conf.
  Comput. Vis.}\hskip 1em plus 0.5em minus 0.4em\relax Springer, 2020, pp.
  536--553.

\bibitem{Chen_2021_ICCV}
Y.~Chen, Z.~Zhang, C.~Yuan, B.~Li, Y.~Deng, and W.~Hu, ``Channel-wise topology
  refinement graph convolution for skeleton-based action recognition,'' in
  \emph{Int. Conf. Comput. Vis.}, October 2021, pp. 13\,359--13\,368.

\bibitem{song2020richly}
Y.-F. Song, Z.~Zhang, C.~Shan, and L.~Wang, ``Richly activated graph
  convolutional network for robust skeleton-based action recognition,''
  \emph{IEEE Trans. Circuit Syst. Video Technol.}, vol.~31, no.~5, pp.
  1915--1925, 2020.

\bibitem{kong2021symmetrical}
J.~Kong, H.~Deng, and M.~Jiang, ``Symmetrical enhanced fusion network for
  skeleton-based action recognition,'' \emph{IEEE Trans. Circuit Syst. Video
  Technol.}, vol.~31, no.~11, pp. 4394--4408, 2021.

\bibitem{zhu2022multilevel}
Y.~Zhu, H.~Shuai, G.~Liu, and Q.~Liu, ``Multilevel spatial--temporal excited
  graph network for skeleton-based action recognition,'' \emph{IEEE Trans.
  Image Process.}, vol.~32, pp. 496--508, 2022.

\bibitem{10113233}
J.~Liu, X.~Wang, C.~Wang, Y.~Gao, and M.~Liu, ``Temporal decoupling graph
  convolutional network for skeleton-based gesture recognition,'' \emph{IEEE
  Trans. Multimedia}, 2023.

\bibitem{xu2023spatiotemporal}
B.~Xu, X.~Shu, J.~Zhang, G.~Dai, and Y.~Song, ``Spatiotemporal
  decouple-and-squeeze contrastive learning for semisupervised skeleton-based
  action recognition,'' \emph{IEEE Trans. Neural Networks Learn. Syst.}, 2023.

\bibitem{aksan2021spatio}
E.~Aksan, M.~Kaufmann, P.~Cao, and O.~Hilliges, ``A spatio-temporal transformer
  for 3d human motion prediction,'' in \emph{3DV}.\hskip 1em plus 0.5em minus
  0.4em\relax IEEE, 2021, pp. 565--574.

\bibitem{yu2023towards}
H.~Yu, X.~Fan, Y.~Hou, W.~Pei, H.~Ge, X.~Yang, D.~Zhou, Q.~Zhang, and M.~Zhang,
  ``Towards realistic 3d human motion prediction with a spatio-temporal
  cross-transformer approach,'' \emph{IEEE Trans. Circuit Syst. Video
  Technol.}, 2023.

\bibitem{shi2020skeleton}
L.~Shi, Y.~Zhang, J.~Cheng, and H.~Lu, ``Skeleton-based action recognition with
  multi-stream adaptive graph convolutional networks,'' \emph{IEEE Trans. Image
  Process.}, vol.~29, pp. 9532--9545, 2020.

\bibitem{zhang2019graph}
X.~Zhang, C.~Xu, X.~Tian, and D.~Tao, ``Graph edge convolutional neural
  networks for skeleton-based action recognition,'' \emph{IEEE Trans. Neural
  Networks Learn. Syst.}, vol.~31, no.~8, pp. 3047--3060, 2019.

\bibitem{cao2018skeleton}
C.~Cao, C.~Lan, Y.~Zhang, W.~Zeng, H.~Lu, and Y.~Zhang, ``Skeleton-based action
  recognition with gated convolutional neural networks,'' \emph{IEEE Trans.
  Circuit Syst. Video Technol.}, vol.~29, no.~11, pp. 3247--3257, 2018.

\bibitem{yang2021feedback}
H.~Yang, D.~Yan, L.~Zhang, Y.~Sun, D.~Li, and S.~J. Maybank, ``Feedback graph
  convolutional network for skeleton-based action recognition,'' \emph{IEEE
  Trans. Image Process.}, vol.~31, pp. 164--175, 2021.

\bibitem{Liu_2020_CVPR}
Z.~Liu, H.~Zhang, Z.~Chen, Z.~Wang, and W.~Ouyang, ``Disentangling and unifying
  graph convolutions for skeleton-based action recognition,'' in \emph{IEEE
  Conf. Comput. Vis. Pattern Recog.}, June 2020.

\bibitem{vaswani2017attention}
A.~Vaswani, N.~Shazeer, N.~Parmar, J.~Uszkoreit, L.~Jones, A.~N. Gomez,
  {\L}.~Kaiser, and I.~Polosukhin, ``Attention is all you need,'' in \emph{Adv.
  Neural Inform. Process. Syst.}, 2017, pp. 5998--6008.

\bibitem{gao2019res2net}
S.-H. Gao, M.-M. Cheng, K.~Zhao, X.-Y. Zhang, M.-H. Yang, and P.~Torr,
  ``Res2net: A new multi-scale backbone architecture,'' \emph{IEEE Trans.
  Pattern Anal. Mach. Intell.}, vol.~43, no.~2, pp. 652--662, 2019.

\bibitem{katharopoulos20a}
A.~Katharopoulos, A.~Vyas, N.~Pappas, and F.~Fleuret, ``Transformers are
  {RNN}s: Fast autoregressive transformers with linear attention,'' in
  \emph{ICML}, ser. Proceedings of Machine Learning Research, H.~D. III and
  A.~Singh, Eds., vol. 119.\hskip 1em plus 0.5em minus 0.4em\relax PMLR, 13--18
  Jul 2020, pp. 5156--5165.

\bibitem{Chen_2020_CVPR}
M.-H. Chen, B.~Li, Y.~Bao, G.~AlRegib, and Z.~Kira, ``Action segmentation with
  joint self-supervised temporal domain adaptation,'' in \emph{IEEE Conf.
  Comput. Vis. Pattern Recog.}, June 2020.

\bibitem{Cao_2017_CVPR}
Z.~Cao, T.~Simon, S.-E. Wei, and Y.~Sheikh, ``Realtime multi-person 2d pose
  estimation using part affinity fields,'' in \emph{IEEE Conf. Comput. Vis.
  Pattern Recog.}, July 2017.

\bibitem{paszke2017automatic}
A.~Paszke, S.~Gross, S.~Chintala, G.~Chanan, E.~Yang, Z.~DeVito, Z.~Lin,
  A.~Desmaison, L.~Antiga, and A.~Lerer, ``Automatic differentiation in
  pytorch,'' \emph{NeurIPSW}, 2017.

\bibitem{paszke2019pytorch}
A.~Paszke, S.~Gross, F.~Massa, A.~Lerer, J.~Bradbury, G.~Chanan, T.~Killeen,
  Z.~Lin, N.~Gimelshein, L.~Antiga \emph{et~al.}, ``Pytorch: An imperative
  style, high-performance deep learning library,'' \emph{Adv. Neural Inform.
  Process. Syst.}, vol.~32, 2019.

\bibitem{carrara2019lstm}
F.~Carrara, P.~Elias, J.~Sedmidubsky, and P.~Zezula, ``Lstm-based real-time
  action detection and prediction in human motion streams,'' \emph{Multimed
  Tools Appl.}, vol.~78, pp. 27\,309--27\,331, 2019.

\bibitem{graves2005framewise}
A.~Graves and J.~Schmidhuber, ``Framewise phoneme classification with
  bidirectional lstm and other neural network architectures,'' \emph{Neural
  networks}, vol.~18, no. 5-6, pp. 602--610, 2005.

\bibitem{van2008visualizing}
L.~Van~der Maaten and G.~Hinton, ``Visualizing data using t-sne.'' \emph{J.
  Mach. Learn. Res.}, vol.~9, no.~11, 2008.

\end{thebibliography}
